\documentclass{article}

\PassOptionsToPackage{numbers, compress}{natbib}
\usepackage[preprint]{neurips_2026}

\usepackage[utf8]{inputenc} % allow utf-8 input
\usepackage[T1]{fontenc}    % use 8-bit T1 fonts
\usepackage{hyperref}       % hyperlinks
\usepackage{url}            % simple URL typesetting
\usepackage{booktabs}       % professional-quality tables
\usepackage{amsfonts}       % blackboard math symbols
\usepackage{nicefrac}       % compact symbols for 1/2, etc.
\usepackage{microtype}      % microtypography
\usepackage{xcolor}         % colors
 \usepackage{graphicx}

 \usepackage{amsmath}
\usepackage{amssymb}
\usepackage{mathtools}
\usepackage{amsthm}
\usepackage{makecell}
\usepackage{nicefrac}
\usepackage{multirow}
\usepackage[table]{xcolor}
\usepackage{amssymb}
\usepackage{enumitem}
\usepackage{placeins}
\usepackage{algorithm}
\usepackage{algorithmic}
\usepackage{wrapfig}
\usepackage{subcaption}
\def\B{\boldsymbol{B}}

\def\K{\boldsymbol{K}} 

\def\O{\boldsymbol{O}}
 
\def\Q{\boldsymbol{Q}} 
\def\R{\mathbb{R}} 
 
\def\S{\boldsymbol{S}}

\def\V{\boldsymbol{V}} 
\def\W{\boldsymbol{W}}

\def\b{\boldsymbol{b}} 

\def\q{\boldsymbol{q}}

\def\v{\boldsymbol{v}} 
 
\def\x{\boldsymbol{x}}

\newcommand{\softmax}{\operatorname{softmax}}

\newcommand{\myparagraph}[1]{\noindent\textbf{#1.}}

\newcommand{\onev}{\boldsymbol{1}}

\newcommand{\ie}{i.e., }
\title{
Mixture-of-Top-$k$ Attention: \\
Efficient Attention via Scalable Fast Weights
}

\author{
  Qishuai Wen, Zhiyuan Huang, Wei He, Xianghan Meng, and Chun-Guang Li\\
  Beijing University of Posts and Telecommunications \\
  \texttt{\{wqs, huangzhiyuan, mengxianghan, wei.he, lichunguang\}@bupt.edu.cn}\\
}

\begin{document}

\maketitle

% \vspace{-10pt}

\begin{abstract}
  The vanilla self-attention mechanism in Transformers can be viewed as a two-layer fast-weight MLP, whose weights are dynamically induced by inputs and whose hidden dimension is equal to the sequence length $N$.
  As the context extends, the expressive capacity of such an $N$-width MLP increases, but it becomes unscalable for extremely long sequences.
  Recently, this fast-weight perspective has motivated the Mixture-of-Experts (MoE) attention mechanism, which partitions the sequence into rigid blocks, treats them as fast-weight experts, and sparsely routes the tokens to them.
  In this paper, we elevate this perspective to a unifying framework for efficient attention mechanisms, interpreting them as making fast weights scalable through either routing or compression, and organizing them into a five-dimensional taxonomy.
  Then, we propose \textbf{Mi}xture-of-\textbf{T}op-$k$ \textbf{A}ttention (\textbf{MiTA}), which employs a small set of landmark queries to gather top-$k$ attended key-value pairs as query-aware and deformable routed experts, while compressing the $N$-width MLP into a narrower shared expert.
  Consequently, our MiTA improves the flexibility of prior MoE attention from rigid to deformable fast-weight experts, as well as the scalability of prior top-$k$ attention from query-specific set to reusable top-$k$ set.
  We conduct extensive experiments on vision tasks showing the superior effectiveness and efficiency of our MiTA, % while 
  and also uncovering intriguing properties such as an emergent token-pruning effect and easy generalization from standard attention.
  Code is available at \href{https://github.com/QishuaiWen/MiTA}{https://github.com/QishuaiWen/MiTA}.
  
\end{abstract}

\section{Introduction}
% \vspace{-10pt}
% 
Attention is the core operation of Transformers, which underpins today's success and wide application of deep learning.
Intuitively, attention learns to store the context as key-value associations and retrieve this short-term memory via queries~\cite{Bietti:NIPS2023}.
However, such an all-to-all lookup paradigm incurs quadratic computation and memory complexity in the sequence length, thereby hindering its scaling to long sequences.
To this end, a plethora of efficient attention methods have been explored, 
% yet without a unifying principle.
albeit still lacking a unifying principle~\cite{zhang2025efficient}.

In recent years, several lines of work have converged on a promising viewpoint: the parameters of a two-layer MLP can be viewed as key-value pairs~\cite{Geva:2021}, and the key-value pairs in full attention can be viewed as the fast (i.e., input-dependent) weights of an MLP~\cite{Schlag:2021, Han:2025-ViTTT}.
Therefore, pursuing an efficient attention can be framed as a \emph{fast-weight scaling} problem, and draw inspiration from slow-weight (i.e., parameter) scaling approaches, such as weight tying~\cite{Wu:2025}, model pruning~\cite{cheng2024survey}, and conditional computation~\cite{Riquelme:2021}. 

Recent work along this direction has demonstrated that sparse attention inspired by Mixture-of-Experts (MoE) can scale up the effective sequence length, deliver wall-clock speedups, and accommodate large-scale pretraining~\cite{Lu:2025-MoBA, Yuan:ACL25-NSA}. 
In contrast to traditional MoE, whose experts are static and structured model parameters (i.e., slow weights), their experts are input-dependent fast weights~\cite{Schmidhuber:1992}, more precisely, subsets of key-value pairs. 
Routing queries to the fast-weight experts sparsely, MoE attention reduces the complexity from quadratic to linear in sequence length $N$.

\begin{table}[t]
  \setlength{\tabcolsep}{5.8pt}
  \caption{\small
  A five-dimensional taxonomy for efficient attention methods from the fast-weight scaling perspective.
  We present expert type and expert count together in the table.
  A detailed analysis of this taxonomy and how MiTA fits into it can be found in Appendix~\ref{sec:fit-tax}.
  }
 \vspace{1pt}
  \label{tab:taxonomy}
  \centering
  \begin{scriptsize}
    \begin{tabular}{lcccc}
    \toprule
    Method  
    & Scaling strategy 
    & Expert type and count 
    & Expert construction
    & Routing topology
    \\
    \midrule
    Linear Attention
    \citeyearpar{katharopoulos2020transformers}
    & \multirow{2}{*}{Compression} & One shared linear layer & Kernelization & All-to-one 
    \\
    MHLA~\cite{zhang2026mhla}
    & & $m$ linear layers 
    & Local prior \& kernelization
    & $m$ times $\frac{N}{m}$-to-one
    \\
    \midrule
    Linformer~\cite{wang2020linformer}
    &
    \multirow{2}{*}{Compression}
    &
    \multirow{2}{*}{One shared MLP}
    &
    \multirow{2}{*}{Learnable projection}
    &
    \multirow{2}{*}{All-to-one}
    \\
    PVT~\cite{Wang2021pvt}
    \\
    \midrule
    Agent Attention~\cite{han2024agent}
    & Compression
    & One shared MLP
    &  
    Landmark probing
    & All-to-one
    \\
    \midrule
    TTT~\cite{sun2024learning}
    & \multirow{2}{*}{Compression} & \multirow{2}{*}{One shared module}
    & \multirow{2}{*}{Test-time training}
    & \multirow{2}{*}{All-to-one}
    \\
    ViT$^3$~\cite{Han:2025-ViTTT} & & & &
    \\
    \midrule
    Deformable DETR~\citeyearpar{zhu2020deformable} 
    & \multirow{2}{*}{Routing}
    & $N$ MLPs
    & \multirow{2}{*}{Offset predicting}
    & $N$ times one-to-one
    \\
    DAT~\cite{xia2022vision}
    & 
    & One shared MLP 
    & 
    & $N$-to-one 
    \\
    \midrule
    BRA~\cite{Zhu:CVPR2023-BRA}
    & Routing
    & \multirow{3}{*}{$\tfrac{N}{n}$ MLPs}
    & \multirow{3}{*}{Locality prior}
    & $\frac{N}{n}$-to-$\frac{N}{n}$
    \\
    MoBA~\cite{Lu:2025-MoBA} 
    & Routing
    & 
    & 
    & \multirow{2}{*}{$N$-to-$\frac{N}{n}$}
    \\
    NSA~\cite{Yuan:ACL25-NSA} 
    & 
    Compression \& Routing
    &
    \\
    \midrule
    Spark Attention~\cite{you2025spark} 
    & \multirow{2}{*}{Routing}
    & \multirow{2}{*}{$N$ MLPs} 
    & \multirow{2}{*}{Light-weight skim}
    & \multirow{2}{*}{$N$ times one-to-one} 
    \\
    DSA~\cite{liu2025deepseek} 
    & & & & 
    \\
    \midrule
    \rowcolor{cyan!10}
    \textbf{MiTA} (ours) 
    & Compression \& Routing 
    & $m$ MLPs
    & Landmark probing
    & $N$-to-$m$
    \\
    \bottomrule
    \end{tabular}
  \end{scriptsize}
  \vspace{-1.5em}
\end{table}

Roughly speaking, the central challenge in scaling the fast weights via MoE is to construct fast-weight experts from unstructured key-value pairs.
For example, given the spatial and temporal locality priors in many modalities, a straightforward, hardware-friendly, albeit naive, approach is to partition the sequence into contiguous, non-overlapping, fixed-size blocks, and regard these blocks as fast-weight experts (i.e., a mixture of blocks or chunks)~\cite{Lu:2025-MoBA, Yuan:ACL25-NSA, Wu:2025-VMoBA, Cai:2025}.
After that, routing vectors are obtained by aggregating each block into a single vector, e.g., via average pooling or parameterized modules.

Since splitting the $N$-width fast-weight MLP into a mixture of such rigid blocks is coarse and suboptimal, subsequent work~\cite{Jia:2025-MoGA} has sought to improve upon the splitting scheme.
Notably, we argue that top-$k$ attention~\cite{you2025spark, liu2025deepseek} can be interpreted as spawning $N$ $k$-width sub-MLPs 
% (\ie experts) 
from the $N$-width MLP, via gathering top-$k$ key-value pairs attended 
% (\ie activated) 
by each query.
Therefore, the rigid (fixed-shape) expert in prior MoE attention methods are replaced by top-$k$ attention 
with top-$k$ key-value pairs.
However, instead of the width of the fast-weight MLP in full attention,
it is the number of such sets of top-$k$ key-value pairs that always equals $N$ in their top-$k$ attention, which still limits its scalability.

\begin{wrapfigure}[18]{r}{0.48\textwidth}
 \centering
  \includegraphics[width=0.48\columnwidth]{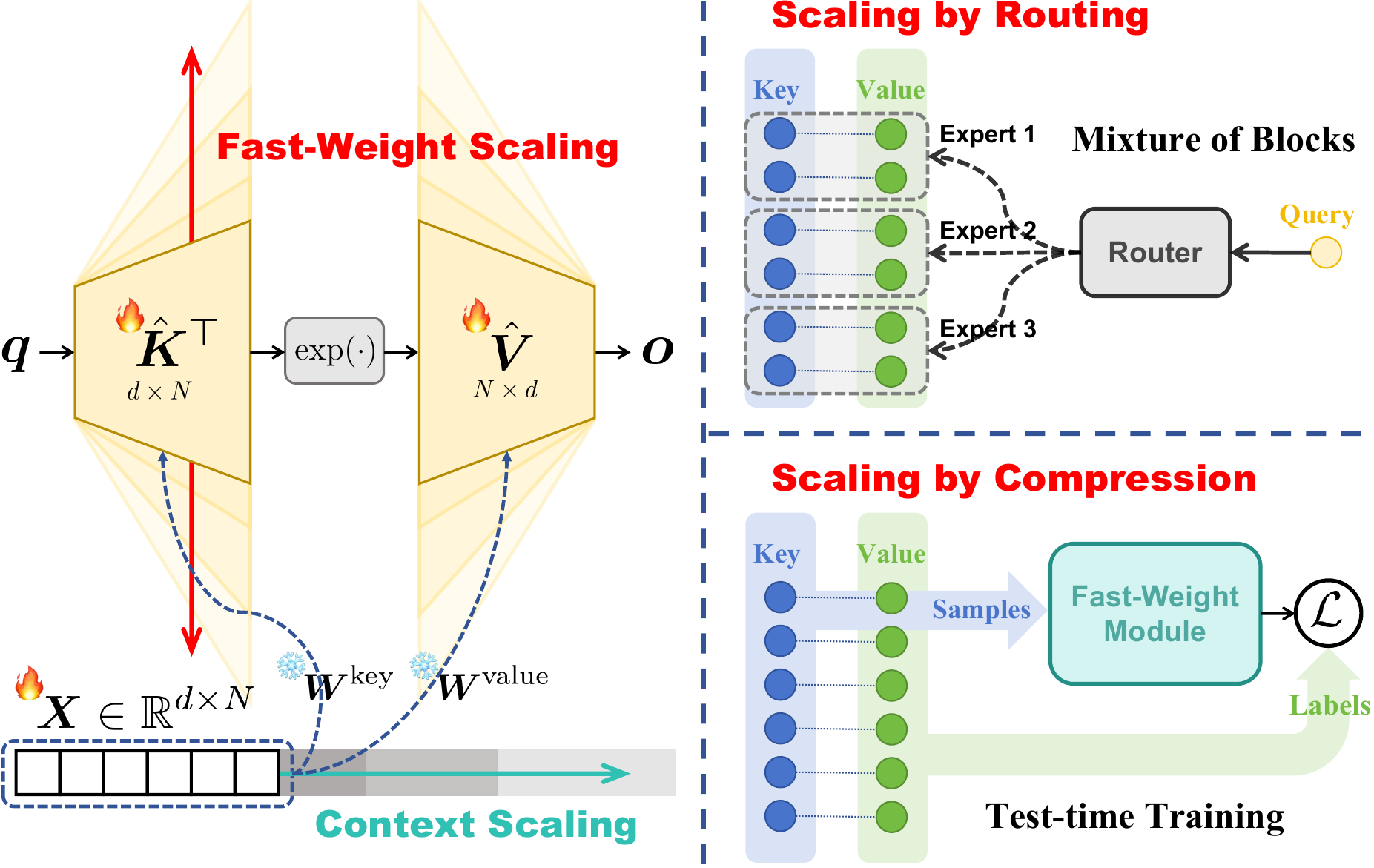}
  \vspace{-1.6em}
  \caption{\footnotesize
  As the context extends, the width of the two-layer fast-weight MLP induced by full attention increases accordingly.
  We categorize efficient fast-weight scaling approaches into two strategies: a) scaling by routing and b) scaling by compression,
  and illustrate each with a representative method: MoBA~\cite{Lu:2025-MoBA} and TTT~\cite{sun2024learning}.
  }
  \label{fig:fast-weight-scaling}
\end{wrapfigure}

On the other hand, unlike the methods mentioned above that scale fast weights by accessing a subset of them, linear attention~\cite{katharopoulos2020transformers} and Test-Time Training (TTT)~\cite{sun2024learning} compress the $N$-width MLP into one (or several~\cite{zhang2026mhla}) light-weight module(s), analogous to model compression~\cite{ba2014deep} and knowledge distillation~\cite{hinton2015distilling}.
Therefore, from a fast-weight scaling perspective, we divide the scaling approaches of existing efficient attention methods discussed in this paper into two general categories: a) scaling by routing, and b) scaling by compression (see Fig.~\ref{fig:fast-weight-scaling} for an illustration).
Scaling by compression alone sacrifices a precise access to the original key-value pairs; whereas scaling by routing alone lacks a global summary of the full context.
Although these two approaches are not mutually exclusive, most existing methods typically adopt only one of them.
% \vspace{-3pt}

% \begin{figure}[h]
%   \centering
%   \includegraphics[width=0.5\columnwidth]{fast_weight_scaling_v5.pdf}
%   \vspace{-5pt}
%   \caption{\small
%   As the context extends, the width of the two-layer fast-weight MLP induced by full attention increases accordingly.
%   % 
%   We categorize efficient fast-weight scaling approaches into two strategies: scaling (a) by routing and (b) by compression,
%   and illustrate each with a representative method: MoBA~\cite{Lu:2025-MoBA} and TTT~\cite{sun2024learning}.
%   }
%   \label{fig:fast-weight-scaling}
%   \vspace{-2em}
% \end{figure}

In this paper, we first elevate the fast-weight scaling perspective into a five-dimensional taxonomy that accommodates a broad spectrum of prior efficient attention methods (see Tab.~\ref{tab:taxonomy}).
While such a taxonomy is by no means complete, it paves the way towards a unified framework and serves as a useful design principle.
Then, we propose \textbf{Mi}xture-of-\textbf{T}op-$k$ \textbf{A}ttention (\textbf{MiTA}), which employs a small set of landmark queries to gather top-$k$ attended key-value pairs as query-aware and deformable routed experts, while compressing the $N$-width MLP into a narrower shared expert.
Consequently, our MiTA improves the flexibility of prior MoE attention from rigid to query-aware and deformable experts, and the scalability of prior top-k attention from $N$ private (i.e., query-specific) to a tunable number $m$ of reusable top-$k$ sets; and b) combines the two scaling strategies, \ie, routing and compression, together, 
% together, thereby addressing the these limitations identified within the taxonomy.
thereby addressing the limitations identified within the taxonomy.

\begin{figure}[htb]
  \vspace{-0.5em}
  \centering
  \includegraphics[width=0.45\columnwidth]{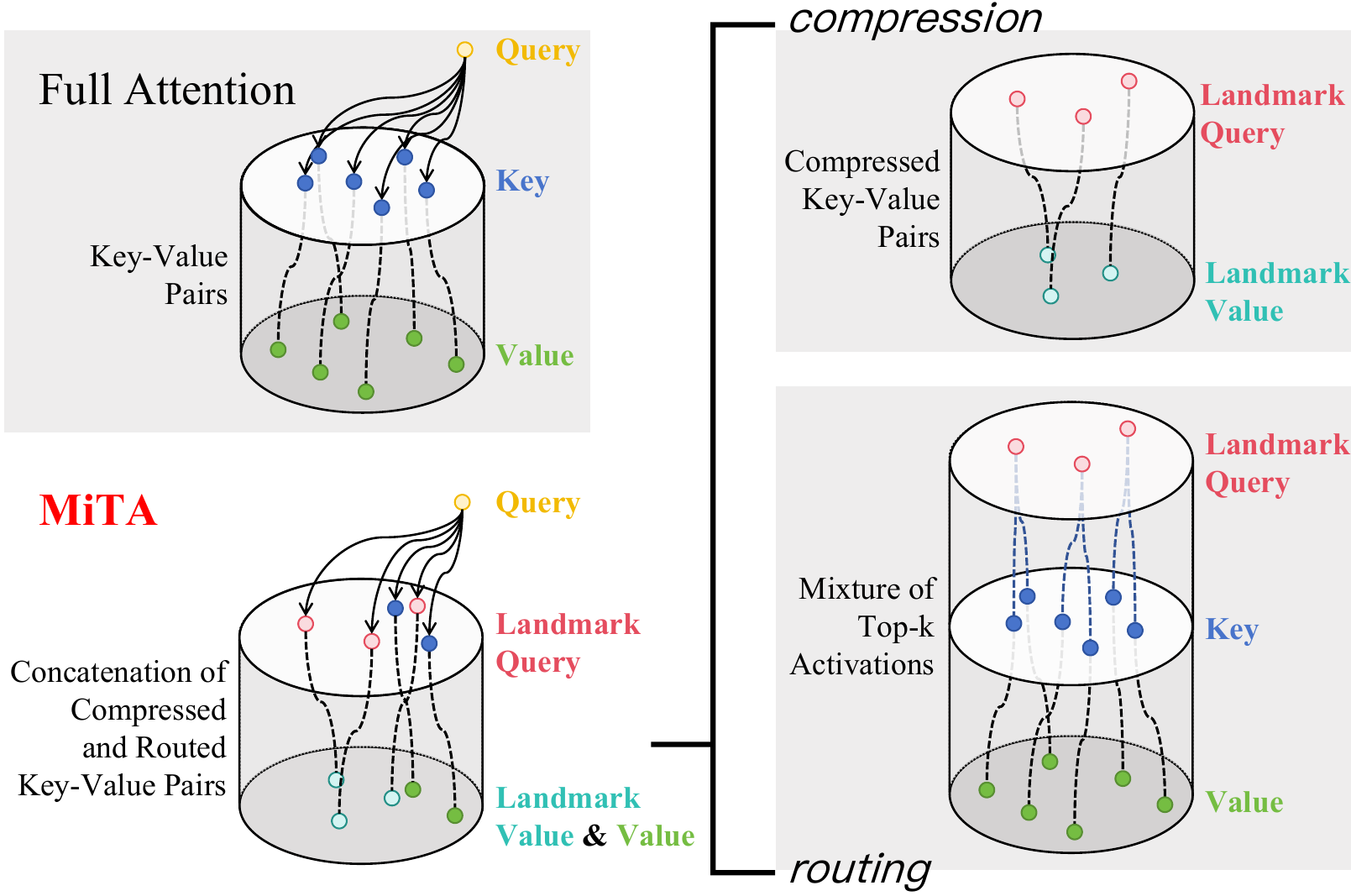}
  \vspace{-1em}
  \vspace{+2pt}
  \caption{\small
  Illustration for our MiTA.
  In full attention,
  each query attends to all key-value pairs. 
  In our MiTA, it attends to 
  the concatenation of 
  a small number (\ie, 3 as shown above) of 
  the compressed key-value pairs 
  and a routed subset of the original key-value pairs (\ie, 2 as shown above).
  }
  \label{fig:mita}
  \vspace{-1em}
\end{figure}

Specifically, our MiTA queries original key-value pairs using landmark queries to obtain corresponding landmark values. 
The resulting landmark query-value pairs (i.e., compressed key-value pairs) form the shared expert, where landmark queries act as keys and landmark values as values.
Meanwhile, MiTA reorganizes the original key-value pairs into routed experts by gathering the top-$k$ key-value pairs attended by each landmark query.
As shown in Fig.~\ref{fig:mita}, for each query,
MiTA concatenates 
these compressed key-value pairs 
with a routed subset of the original key-value pairs.

% \begin{wrapfigure}[15]{r}{0.48\textwidth}
% % \begin{figure}[htb]
%   \vspace{-0.5em}
%   \centering
%   \includegraphics[width=0.48\columnwidth]{mita_v9.pdf}
%   \vspace{-1em}
%   \vspace{-5pt}
%   \caption{\small
%   Illustration for our MiTA.
%   In full attention,
%   each query attends to all key-value pairs. 
%   % 
%   In our MiTA, it attends to 
%   the concatenation of 
%   a small number (e.g., 3 as shown above) of 
%   the compressed key-value pairs 
%   and a routed subset of the original key-value pairs (2 above).
%   }
%   \label{fig:mita}
%   \vspace{-1em}
% % \end{figure}
% \end{wrapfigure}

% \vspace{-8pt}
%{\mcr 
\paragraph{Contributions.} 
% Contributions of the paper are summarized as follows.
% Contributions of the paper can be summarized into three-fold as follows.
The contributions of this paper are three-fold:
\begin{enumerate}[leftmargin=*,topsep=0.25em,noitemsep]
  \item 
  % Conceptually, we introduce a unified perspective that views efficient attention mechanisms as scaling fast weights, systematically organizing existing methods into a five-dimensional taxonomy and naturally revealing their limitations: the rigidity of experts, the non-reusability of top-$k$ set, and the necessity of combining routing and compression.
  \textbf{Conceptually}, we introduce a unified perspective that views efficient attention mechanisms as scaling fast weights, systematically organizing existing methods into a five-dimensional taxonomy and naturally revealing their limitations: the rigidity of experts, the non-reusability of top-$k$ sets, and the need to combine routing and compression.
  \item 
  % Technically, we propose a novel efficient attention, MiTA, which construct a tunable number of query-aware, deformable experts (or reusable, routable top-$k$ set) and achieves scalable fast weights through both routing and compression, thereby addressing these limitations in a principled manner.
  \textbf{Technically}, we propose a 
  % simple yet effective 
  novel
  efficient attention mechanism, MiTA, which constructs a tunable number of query-aware, deformable expert (equivalently, reusable and routable top-$k$ sets) and achieves scalable fast weights through both routing and compression, thereby addressing these limitations in a principled manner.
  \item 
  % Experimentally, we validate the effectiveness and efficiency of MiTA on vision tasks, where it surpasses strong baselines by a large margin while using less wall-clock time.
  \textbf{Experimentally}, we validate the effectiveness and efficiency of MiTA on vision tasks, %where it 
  demonstrating that our MiTA surpasses strong baselines by a large margin while using less wall-clock time, efficiently approaching the expressive capacity of full attention to a remarkable extent.
\end{enumerate}

% \begin{wrapfigure}[13]{r}{0.45\textwidth}
% \end{wrapfigure}

\section{Related Work} \label{sec:related-work}

% Although our work primarily focuses on the vision domain, we provide a high-level and comprehensive discussion of related work across different modalities.
% % 
% We hope this helps demystify the underlying principles of efficient attention.

\subsection{Fast-Weight Scaling}
In contrast to slow weights (i.e., model parameters that remain unchanged after training and encode persistent knowledge), 
% encoding persistent knowledge,
fast weights are input-conditioned and act as temporary parameters.
They can be viewed as short-term memory~\cite{behrouz2024titans}, playing an important role in meta-learning~\cite{kirsch2022general}.
% and in-context learning~\cite{chan2022data}.
% 
While most scaling efforts have focused on expanding slow weights (e.g., increasing model width and depth)~\cite{kaplan2020scaling, tan2019efficientnet},
extending Transformers' sequence length turns out to implicitly scale fast weights.

For instance, Test-Time Training (TTT)~\cite{sun2024learning} effectively
% generally
compresses the $N$-width two-layer fast-weight MLP in full attention into a smaller fast-weight module, likewise for linear attention~\cite{katharopoulos2020transformers} into a linear layer~\cite{Han:2025-ViTTT} and Linformer~\cite{wang2020linformer} into a narrower MLP.
Rather than scaling by compressing the full fast-weight MLP, MoE attention factorizes it into experts and access them sparsely via routing.

% In our MiTA, the two strategies mentioned above
% ---scaling by compressing and scaling by sparse routing---
% are combined.
Our MiTA combines the two strategies mentioned above.
% 
% It follows the Agent Attention~\cite{han2024agent} to 
It compresses full key-value pairs into a smaller set (a narrower MLP), which offers a coarse yet global summary, and then leverages the top-$k$ activations of the 
% agent tokens (i.e., 
landmark queries
% ) 
to identify underlying experts, which enables precise retrieval.

\subsection{MoE-Inspired Sparse Attention}
Previous sparse attention methods typically focus on the design of fixed sparse patterns (e.g., the local window, axial stripe~\cite{Dong:CVPR2022-CSWin}, and the vertical-slash pattern~\cite{jiang2024minference}) or learnable ones (e.g., hash buckets~\cite{kitaev2020reformer} and $k$-means clusters~\cite{roy2021efficient}.)
To maintain a global connectivity, shared memory (i.e., tokens that attend to and are attended by all tokens) is also introduced~\cite{beltagy2020longformer}.
However, these patterns are either task-dependent~\cite{lai2025flexprefill} or too burdensome to achieve wall-clock speedup~\cite{dao2022flashattention}.

Recently, sparse attention inspired by Mixture-of-Experts (MoE) has been applied to pre-training large language models, yielding extended effective context with reduced overhead~\cite{Lu:2025-MoBA, Yuan:ACL25-NSA}.
Its key advantage is the routing mechanism, which enables query-aware selection within the key-value cache.
Consequently, despite coarse selection granularity (e.g., evenly split, non-overlapping blocks), the sparse pattern is highly flexible and can vary across queries.

In this paper, we argue that, beyond routing, the experts can also be query-aware and deformable.
Specifically, for an arbitrarily long sequence, we can build a fixed, yet configurable number of experts by gathering semantically related key-value pairs.
A similarly motivated line of research is to replace the fixed, non-overlapping rectangular patchification in ViTs~\cite{dosovitskiy2020image} with a deformable, content-adaptive tokenization scheme~\cite{chen2021dpt}.

\subsection{Deformable Attention}
More broadly, conditional computation, the idea underlying MoE, has been explored through the deformable convolution~\cite{dai2017deformable} in convolutional neural networks.
And deformable attention was introduced for object detection by \citet{zhu2020deformable} and later generalized to Vision Transformers by \citet{xia2022vision}.
Given a query, while being blind to keys, deformable attention predicts offsets relative to a small set of default positions of the keys, thereby inducing a query-aware, deformable sparse attention pattern (\ie a fast-weight expert in our context).

% Thanks to the 
Recent top-$k$ attention methods, e.g., Spark Attention~\cite{you2025spark} and DeepSeek Sparse Attention~\cite{liu2025deepseek}, which have been applied to the training of Gemma 3n and DeepSeek-V3.2, respectively, can also be viewed as improved deformable attention.
Instead of predicting the spatial positions solely from the query as in prior deformable attention, these methods locate the top-$k$ key-value pairs by allowing each query to take a lightweight skim of the full keys, e.g., using partial features~\cite{you2025spark} or low precision (FP8)~\cite{liu2025deepseek}.

Our MiTA retains the key advantage of top-$k$ attention by constructing deformable experts (i.e., sparse patterns) that depend on both queries and keys.
However, unlike the methods discussed above, we replace the per-query offset predicting~\cite{zhu2020deformable, xia2022vision} or light-weight skim~\cite{you2025spark, liu2025deepseek} with
more efficient per-query routing to reusable sparse patterns, thereby further improving scalability.

\section{Methods}

\subsection{Efficient Attention as Scaling Fast Weights} \label{sec:fast-weight-scaling}
% 
%In this 
This subsection will first present 
the mathematical formulation of 
the fast-weight MLP 
equivalent to full attention,
% that full attention is equivalent to,
% in full attention, 
and then reinterpret efficient attention 
as a fast-weight scaling problem. 
From this perspective, 
%we propose 
a taxonomy of previous efficient attention methods 
in terms of how they scale fast weights will be given.

\myparagraph{Fast-weight MLP}
Formally, 
full attention (i.e., scaled dot-product attention~\cite{vaswani2017attention})
% ; SDPA)
can be written as:
\begin{align}
    \operatorname{Atten}(\q, \K, \V) = \V \softmax\left(\K^\top \q / \sqrt{d}\right),
    \label{eq:full-attn}
\end{align}
where $\q \in \R^d$ is a query,
and the columns of $\K, \V \in \R^{d\times N}$ 
are the keys and values, respectively.
By contrast, a $D$-width two-layer MLP,
i.e., the feed-forward network (FFN) in Transformers, 
is: 
\begin{align}
    \operatorname{MLP}_{\sigma}(\x \mid \W_1; \W_2) = 
    \W_2 \sigma\left(\W_1^\top \x + \b_1\right) + \b_2,
    \label{eq:MLP}
\end{align}
where $\x \in \R^d$ is the input vector,
$\W_1, \W_2 \in \R^{d\times D}$ are weights,
$\b_1 \in \R^{D}$ and $\b_2 \in \R^{d}$ are biases\footnote{
For notion %al 
simplicity, %the 
biases are omitted from the argument. % list.
},
% which are omitted from the parameter list for notional simplicity,
and $\sigma(\cdot)$ is an element-wise activation function (e.g., ReLU).
Note that full attention in Eq.~\eqref{eq:full-attn} is equivalent to
the following MLP:
\begin{align}
    % \operatorname{SDPA}(\q, \K, \V) =
    \operatorname{MLP}_{\exp}(\q \mid \hat{\K}; \hat{\V})
    = \hat{\V}\exp\left(\hat{\K}^\top \q \right), \label{eq:fast-weight-MLP} \\
    \hat{\K} = \K / \sqrt{d},\ 
    \hat{\V} = \V / 
    \left(
    \exp(\hat{\K}^\top \q)^\top \onev_N
    \right),
\end{align}
where the weights are given by the 
scaled keys and values,
the biases are all zero,
and the activation function is the exponential function $\exp(\cdot)$.
Therefore, we have demonstrated that 
full attention in Eq.~\eqref{eq:full-attn} is equivalent to 
the \emph{$N$-width, fast-weight, two-layer MLP} in Eq.~\eqref{eq:fast-weight-MLP}.

\myparagraph{Unbounded fast-weight scaling}
Because the hidden dimension of 
the fast-weight MLP in Eq.~\eqref{eq:fast-weight-MLP}
always equals to the sequence length $N$,
the per-query overhead grows linearly with $N$ rather than 
fixed as in the slow-weight MLP in Eq.~\eqref{eq:MLP}.
Thus, under the all-to-all lookup paradigm, processing $N$ queries incurs an overall $\mathcal{O}(N^2)$ overhead. 
This unbounded, rapid growth prevents scaling the fast weights to arbitrarily long sequences. 
While this challenge is commonly described as the quadratic complexity of full attention 
in the literature on efficient attention,
% efficient attention literature,
we refer to it as the \emph{unbounded fast-weight scaling} issue.

\myparagraph{Fast-weight scaling taxonomy}
% \myparagraph{A taxonomy of fast-weight scaling}
% 
Although we have discussed relevant efficient-attention methods in Sec.~\ref{sec:related-work},
they have not been systematically organized %under 
into a unified fast-weight scaling perspective.
We therefore propose a five-dimensional taxonomy:
% with the following dimensions: 
a) \emph{scaling strategy}, i.e., 
scaling the $N$-width fast-weight MLP either by compressing it into a 
% narrower 
light-weight
module or by routing each query to a subset of it;
b) \emph{expert 
% count and type
% type
count},
% and (c) \emph{expert count}
i.e., 
how many fast-weight experts are constructed (in particular, scaling by compression typically yields a single shared expert);
c) \emph{expert type}, i.e.,
what module each expert takes (e.g., an MLP or a linear layer);
d) \emph{expert construction}, i.e., how experts are formed from the key-value pairs; 
and
e) \emph{routing topology}, i.e., the query-expert assignment pattern.
The mapping of existing %prior 
methods to this taxonomy is summarized in Tab.~\ref{tab:taxonomy}.
We defer  
the in-depth 
% a detailed xc
discussion until %after presenting our method (i.e., to 
Appendix~\ref{sec:fit-tax}. %).

\subsection{Mixture-of-Top-$k$ Attention (MiTA)} \label{sec:mita}

Motivated by the limitations identified through our fast-weight scaling perspective, our method has two design goals: 
a) moving from rigid to query-aware, deformable experts, 
and from private (i.e., query-specific) to reusable top-$k$ key-value pairs, and 
b) combining compression and routing. 
% 
% We start from the second goal, which naturally leads to the first one.
We begin with the second goal, which is simpler, and show that it naturally leads to the first, which is more important.

Note that both scaling by compression and scaling by routing have the inherent limitations.
Compressing the full fast-weight MLP inevitably discards information, 
although the information loss can be mitigated by adopting milder compression schemes 
(e.g., piecewise compression~\cite{zhang2026mhla}) 
or by compressing into a more expressive module (e.g., via test-time training~\cite{sun2024learning}). 
In contrast, scaling by routing is faithful and lossless, but it often lacks a global view of the context.

\begin{wrapfigure}[18]{r}{0.48\textwidth}
\vspace{-12pt}
% \begin{figure}[h]
  \centering
  \includegraphics[width=0.46\columnwidth]{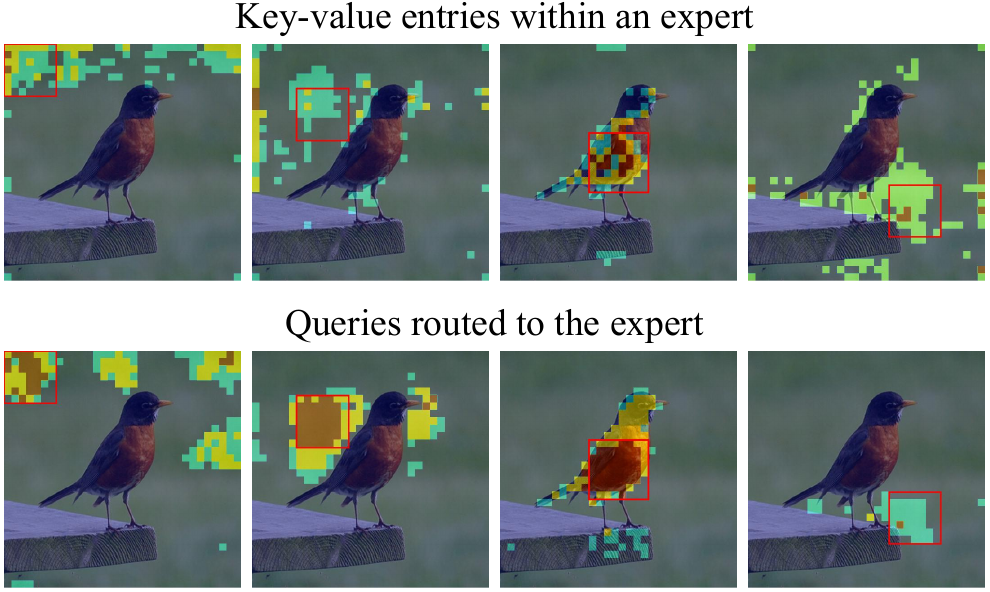}
  \vspace{-0.55em}
  \caption{\small
  Visualization of
  experts' 
  gathered
  % landmark query,
  key-value pairs, and
  routed queries.
  The red box marks the local window from which the landmark query is obtained via average pooling.
  The attention heatmap (averaged over attention heads) indicates 
  key-value pairs within each expert (top row) and the queries routed to it (bottom row).
  Notably, neither the expert’s key–value pairs nor the routed queries are confined to the local window.
  }
  \label{fig:show-expert}
  % \vspace{-pt}
% \end{figure}
\end{wrapfigure}

To combine the strengths of both strategies, 
i.e., to retain a compact global summary while enabling precise, token-level retrieval,
we introduce a small set of \emph{landmark queries} $\tilde{\Q}\in\R^{d\times m}$ with $m \ll N$. 
These landmark queries probe the full key-value cache 
and then, via cross-attention, 
compress it into a global fast-weight module, 
while simultaneously forming deformable fast-weight experts by gathering the top-$k$ activated (i.e., attended) key--value pairs for each landmark query. 

This design is motivated by two observations: 
a) register tokens can attend to distinct semantic regions of an image~\cite{darcet2023vision}; 
and b) class embeddings~\cite{strudel2021segmenter} (or object queries~\cite{carion2020end} and mask embeddings~\cite{cheng2021per}) provide a compact global summary for dense prediction tasks~\cite{wen2024rethinking} and even image generation~\cite{yu2024image}.
% 
% we employ a small set of landmark queries $\tilde{\Q}\in\R^{d\times m}$, where $m \ll N$,
% to look up the full key--value cache,
% and gather, for each landmark query, 
% the top-$k$ 
% % attended 
% activated
% key--value entries
% to construct a fast-weight expert.
% % 
% 
The landmark queries can be obtained
in various ways, 
e.g., by assigning a set of learnable slow weights
or by downsampling the sequence using
% cross-attention or 
a convolutional module. 
In this work, we simply apply average pooling 
over uniformly spaced, equal-sized windows, 
as shown in Fig.~\ref{fig:show-expert}.

Specifically, 
% given a routed query $\q$, 
the $i$-th landmark query $\tilde{\q}_i$ (among $m$ in total) defines the $i$-th expert $\mathcal{E}_i$ as follows: 
\begin{align}
    \mathcal{E}_i(\q) &= 
    \operatorname{Atten}(\q, \K^{(i)}, \V^{(i)}),\ i \in \{1, \ldots, m\},
    \label{eq:mita-expert}
    \\
    \K^{(i)} &= \K_{:,\mathcal{I}_i},\ 
    \V^{(i)} = \V_{:,\mathcal{I}_i} \in \R^{d\times k},
    \label{eq:gather}
    \\
    \mathcal{I}_i &= \operatorname{Top}_k\left(\K^\top \tilde{\q}_i\right) \in \{1, \ldots, N\}^k,
    \label{eq:top-k}
    % \s_i = \K^\top \tilde{\q}_i \in \R^n,
\end{align}
% 
% {\allowdisplaybreaks[2]
% \begin{align}
%     \mathcal{E}_i(\q) = 
%     \operatorname{SDPA}(\q, \K^{(i)}, \V^{(i)}),\ i \in \{1, \ldots, m\},
%     \label{eq:mita-expert}
%     \\ \displaybreak[2]
%     \K^{(i)} = \K_{:,\mathcal{I}_i},\ 
%     \V^{(i)} = \V_{:,\mathcal{I}_i} \in \R^{d\times k},
%     \label{eq:gather}
%     \\ \displaybreak[2]
%     \mathcal{I}_i = \operatorname{Top}_k\left(\K^\top \tilde{\q}_i\right) \in \{1, \ldots, n\}^k,
%     \label{eq:top-k}
% \end{align}}
%
where $\q$ is a query routed to expert $\mathcal{E}_i$, and
$\K^{(i)}, \V^{(i)}$ are the top-$k$ key-value pairs activated by $\tilde{\q}_i$.
% 
% Moreover, 
This approach effectively reorganizes
the full key-value pairs into a mixture of top-$k$ key-value pairs
(i.e., a mixture of deformable experts).
We directly use 
$\tilde{\q}_i$ as the routing vector for expert $\mathcal{E}_i$.
Then the routing logits from the $N$ queries $\Q \in \R^{d\times N}$ to the $m$ experts
is
$\Q^\top\tilde{\Q} \in \R^{N\times m}$.
% \begin{align}
%     \mathcal{A} = \Q^\top\tilde{\Q} \in \R^{N\times m}
% \end{align}
% 
We denote 
the index of the rank-$j$ expert 
that a query $\q$ is routed to 
under these logits as $e_j(\q) \in \{1, \ldots, m\}$.

Moreover,
regarding the compressed global module, which can be viewed as 
% to create 
a shared expert~\cite{dai2024deepseekmoe}, 
we extract a set of landmark values $\tilde{\V} \in \R^{d\times m}$ via cross-attention
using the landmark queries. 
The $i$-th landmark value corresponding to the $i$-th landmark query is given by:
\begin{align}
    \tilde{\v}_i = \operatorname{Atten}(\tilde{\q}_i, \K, \V). \label{eq:landmark-value}
\end{align}
Then, the shared expert $\tilde{\mathcal{E}}$ is defined as:
\begin{align}
    \tilde{\mathcal{E}}(\q) = \operatorname{Atten}(\q, \tilde{\Q}, \tilde{\V}), \label{eq:shared-expert}
\end{align}
where landmark queries act as the keys in standard attention.
Note that the computations in Eqs.~\eqref{eq:top-k} and \eqref{eq:landmark-value} can be
largely shared, 
where landmark queries are both querying the keys.
Likewise, Eq.~\eqref{eq:shared-expert} can reuse the routing logits.
% Therefore, 
% In other word,
% the computation of the two scaling strategies are closely coupled in our \textbf{MiTA}. 
Therefore, the computations of the two scaling strategies are tightly coupled in our MiTA.

Rather than a straightforward MoE implementation, 
which isolates the computation within each expert 
and aggregates their outputs 
via a weighted sum, 
we concatenate the experts as a single standard attention,
as shown in Fig.~\ref{fig:mita}.
Therefore, our
% \textbf{Mi}xture of \textbf{T}op-k \textbf{A}ctivation (\textbf{MiTA}) 
\textbf{MiTA}
can be written as:
\begin{align}
    \operatorname{MiTA}(\q) &=
    \V^\star \softmax({\K^\star}^\top \q / \sqrt{d}), \label{eq:mita} \\
    \K^\star &= [\tilde{\Q}, \K^{(e_1(\q))}, \ldots, \K^{(e_s(\q))}], \\
    \V^\star &= [\tilde{\V}, \V^{(e_1(\q))}, \ldots, \V^{(e_s(\q))}],
\end{align}
where $\K^\star, \V^\star \in \R^{d\times (m+ks)}$ are the key-value pairs that a query $\q$ attends to in our MiTA, 
and $s$ is the number of 
% selected 
routed
experts (not counting the shared expert) per query.
% \footnote{$m$ is the number of routed expert in total.}
Note that, in practice, the attention can still be computed expert-wise and then combined, 
thanks to the online softmax~\cite{milakov2018online}, as in the FlashAttention~\cite{dao2023flashattention} implementation of \citet{Lu:2025-MoBA}.

\myparagraph{Implementations}
% MiTA is a conceptually simple method, but is effective and technically non-trivial to achive wall-clock speedup.
MiTA is conceptually simple yet effective, but technically non-trivial to realize with actual wall-clock speedups.
 % and complexity analysis
In particular, 
we implement our MiTA in Eq.~\eqref{eq:mita} with $s=1$ (as in Llama 4), 
i.e., each query is fed to the shared expert and routed to
exactly one additional fast-weight expert.
% via quicksort.
% , as is the case in Llama 4.
% 
The %corresponding 
pseudocode of our MiTA is given in Algorithm~\ref{alg:mita} (Appendix B) 
which matches queries to experts by sorting queries according to their expert assignments $e_1(\q)$.
The more general but more complex
case with $s>1$ can be implemented %realized using the MoBA implementation
via MoBA~\cite{Lu:2025-MoBA} at the cost of reduced speed.

\myparagraph{Complexity analysis}
% 
% For Eq.~\eqref{eq:mita} alone, 
The computational complexity of Eq.~\eqref{eq:mita} alone is $\mathcal{O}(N(m + ks))$,
whereas full attention incurs $\mathcal{O}(N^2)$; in practice, $N \gg m+ks$.
Moreover, 
the cross-attention
% -like 
operations in Eqs.~\eqref{eq:landmark-value} and~\eqref{eq:mita}
can be accelerated by the FlashAttention family~\cite{dao2022flashattention, dao2023flashattention, shah2024flashattention}.
The primary bottleneck comes from the gather operation in Eq.~\eqref{eq:gather}:
it entails irregular (random) memory access 
and is therefore likely to make the overall pipeline memory-bound.
Nonetheless, 
% prior top-$k$ attention 
Deepseek Sparse Attention
% has demonstrated that 
suggests that 
this limitation can be mitigated via its optimized implementation~\citep[Fig.~3]{liu2025deepseek}.
Moreover, 
compared with prior top-$k$ attention~\cite{you2025spark, liu2025deepseek}, 
which instantiate a private fast-weight expert per query
($N$ experts in total) 
without routing, 
our MiTA uses a 
% fixed 
fixed (yet tunable)
number (\ie $m$) of fast-weight experts, which is more hardware-friendly, 
at the cost of an $N$-to-$m$ routing step, which is
% also
explored in MoBA~\cite{Lu:2025-MoBA}.
In summary, 
the hardware operations required by MiTA have been validated by prior engineering-centric work, allowing us to focus on the fast-weight scaling perspective, its taxonomy, the construction of a tunable number of query-aware, deformable fast-weight experts, and the combination of compression and routing.

% \clearpage

\begin{figure}[t!]
  \centering
  \includegraphics[width=0.95\textwidth]{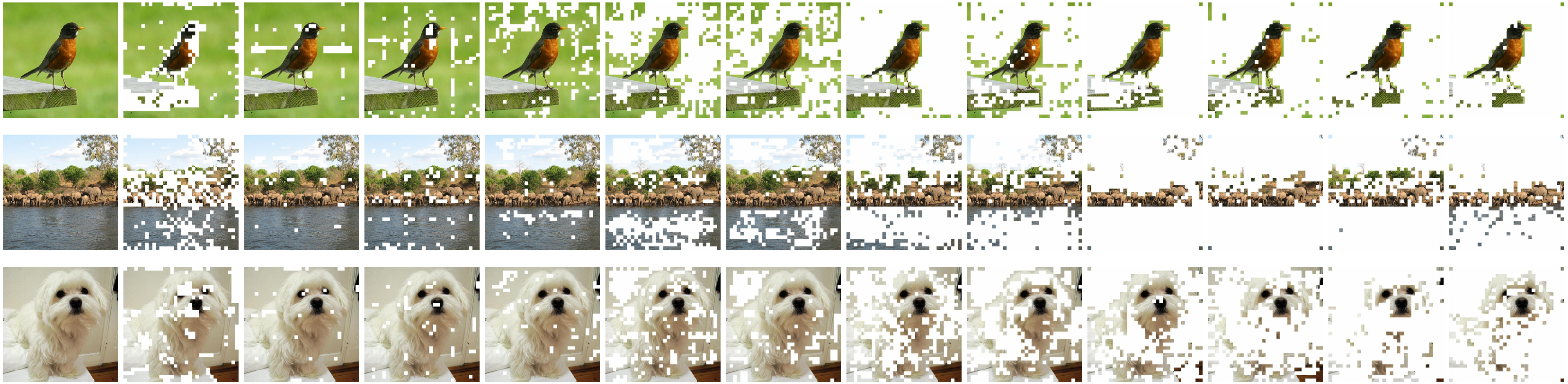}
  \vspace{-0.6em}
  % \vspace{-5pt}
  \caption{\small
  The token pruning effect of MiTA.
  Each row visualizes, for each layer, the positions of key-value pairs (aggregated over heads) selected as experts;
  the leftmost image shows the original input.
  In later layers, most tokens are effectively ``pruned'' (i.e., not selected as experts), and attention concentrates on class-relevant regions. 
  The examples are sampled from the ImageNet-1K training set.
  }
  \vspace{-1em}
  \label{fig:token-pruning}
\end{figure}

\section{Experiments}
To verify the efficacy of our 
% proposed
MiTA, %method, 
we conduct image classification experiments on ImageNet-1K~\cite{deng2009imagenet} 
and semantic segmentation experiments on ADE20K~\cite{zhou2019semantic}.
Moreover, % Additionally, 
we assess the long sequence modeling capability of MiTA on the Long Range Arena benchmark~\cite{tay2020long},
and report its wall-clock
% training and 
training/inference speed against full attention on extremely long sequences.
A detailed ablation study
is provided in Appendix~\ref{sec:ablation}.
% 
% Furthermore, 
% we study the robustness 
% % (more precisely, generalization) 
% (or generalization)
% of MiTA 
% under varying expert count ($m$) and width ($k$) %changes
% when the model is fixed after training.
% % 
% Additionally, we explore a broader problem of interest: the \emph{algorithmic generalization} of Transformers across attention mechanisms.
% 
Furthermore, we study the robustness (or generalization) of MiTA under varying 
expert count ($m$) and width ($k$) with a fixed trained model, and examine the broader algorithmic generalization of Transformers across attention mechanisms; see Appendix~\ref{sec:generalization} for details.

{
\setlength{\floatsep}{0.0em} 
\setlength{\intextsep}{0.0em}

\begin{table}[h]
  \caption{\small
  Results on ImageNet-1K training from scratch. 
  % For a fair comparison, we train all models using the DeiT recipe, with the attention mechanism being the only difference throughout the entire process.
  % 
  $^{\mathrm{Bias}}$: using the agent bias~\cite{han2024agent}; $^{\mathrm{DWC}}$: using depth-wise convolution; $^\mathrm{Gate}$: adding the attention output to the residual with data-dependent gating.
  PolaFormer~\cite{mengpolaformer}, MHLA~\cite{zhang2026mhla}, and FLatten~\cite{han2023flatten} are variants of linear attention. 
  % 
  % Bi-level Routing Attention (BRA)
  BRA~\cite{Zhu:CVPR2023-BRA} and Agent Attention~\cite{han2024agent} can be viewed as degenerate cases of MiTA, scaling by only routing and only compression, respectively.
  }
  % \vspace{-5pt}
  \centering
  \begin{footnotesize} 
  \begin{tabular}{lccc}
    \toprule
    Methods  
    & \# Params
    &  FLOPs  % \footnote{Strictly speaking, the reported FLOPs are MACs}
    &  Acc. (\%)
    \\
    \midrule
    DeiT-T
    % ~(Baseline) 
    & 5.7M & 1.2G & \cellcolor{gray!15}72.2 \\
    Linear-DeiT-T & 5.7M & 1.1G & 68.0 \\
    PolaFormer-DeiT-T & 5.7M & 1.2G & 68.1 \\
    MHLA-DeiT-T & 5.7M & 1.1G & 69.2 \\
    FLatten-DeiT-T & 6.1M & 1.1G & 70.2 \\
    BRA-DeiT-T & 5.7M & 1.1G & 70.2 \\
    Agent-DeiT-T & 5.7M & 1.1G & 70.3 \\
    Agent-DeiT-T$^{\mathrm{Bias}}$ & 6.0M & 1.2G & 71.1 \\
    % \rowcolor{gray!15} 
    \rowcolor{cyan!10}
    MiTA-DeiT-T & 5.7M & 1.1G & 71.1{\scriptsize~(-1.1)} \\
    % \rowcolor{gray!15}
    \rowcolor{cyan!10}
    MiTA-DeiT-T$^{\mathrm{DWC}}$ & 5.7M & 1.1G & 73.4{\scriptsize~(+1.2)} \\
    \midrule
    DeiT-S
    % ~(\textbf{Baseline}) 
    & 22M & 4.6G & \cellcolor{gray!15}79.9 \\
    Agent-DeiT-S & 22M & 4.4G & 78.5 \\
    Agent-DeiT-S$^{\mathrm{Bias}}$ & 23M & 4.4G & 79.1 \\
    % \rowcolor{gray!15} 
    \rowcolor{cyan!10}
    MiTA-DeiT-S & 22M & 4.4G & 79.8{\scriptsize~(-0.1)}\\
    % \rowcolor{gray!15} 
    \rowcolor{cyan!10}
    MiTA-DeiT-S$^{\mathrm{DWC}}$ & 22M & 4.4G & 80.6{\scriptsize~(+0.7)} \\
    % \rowcolor{gray!15} 
    \rowcolor{cyan!10}
    MiTA-DeiT-S$^{\mathrm{DWC}, \mathrm{Gate}}$ & 22M & 4.7G & 81.2{\scriptsize~(+1.3)} \\
    \bottomrule
  \end{tabular}
  \end{footnotesize}
  \label{tab:in1k-fair}
  % \vspace{-2em}
  % \vspace{+5pt}
\end{table}

\vspace{+5pt}

\begin{table}[h]
  \caption{\small
  Comparison with SOTA and efficient ViTs on ImageNet-1K.
  We compare against a recent state-of-the-art ViT variant, ViT-5~\cite{wang2026vit-5}, 
  as well as strong baselines such as DeiT and DeiT-III~\cite{touvron2022deit-iii}. 
  We also include efficient ViTs, including sparse models (SViTE~\cite{chen2021chasing} and Sparsifiner~\cite{wei2023sparsifiner}) 
  and scaling-by-compression models (InLine~\cite{han2024bridging} and Agent Attention).
  }
  % \vspace{-5pt}
  \centering
  \begin{footnotesize}
  \begin{tabular}{lccc}
    \toprule
    Methods  
    & \# Params
    &  FLOPs
    &  Acc. (\%)
    \\
    \midrule
    ViT-5-S
    % ~(\textbf{Baseline}) 
    & 22M & 4.7G & \cellcolor{gray!15}82.2 \\
    DeiT-S & 22M & 4.6G & 79.9 \\
    DeiT-III-S & 22M & 4.6G & 81.4 \\
    SViTE-S & 11M & 2.5G & 79.7 \\
    Sparsifiner-S & 23M & 4.5G & 79.9 \\
    InLine-DeiT-S & 17M & 5.0G & 80.2 \\
    Agent-DeiT-S$^{\mathrm{Bias},\mathrm{DWC}}$ & 23M & 4.4G & 80.5 \\
    MHLA-DeiT-S$^{\mathrm{DWC}}$ & 22M & 4.2G & 81.0 \\
    % \rowcolor{gray!15} 
    \rowcolor{cyan!10}
    MiTA-ViT-5-S & 22M & 4.5G & 81.3{\scriptsize~(-0.9)} \\
    % \rowcolor{gray!15} 
    \rowcolor{cyan!10}  
    MiTA-ViT-5-S$^{\mathrm{DWC}}$ & 22M & 4.5G & 81.7{\scriptsize~(-0.5)} \\
    \bottomrule
  \end{tabular}
  \end{footnotesize}
  \label{tab:in1k-sota}
  % \vspace{-2em}
\end{table}

% \vspace{+5pt}

\begin{table}[t]
  \setlength{\tabcolsep}{2pt}
  \caption{\small
  Results on ADE20K semantic segmentation.
  $^{\triangledown}$: the attention mechanism is directly replaced, rather than being natively pretrained.
  Mask Transformer is the decoder introduced in the pioneering work Segmenter~\cite{strudel2021segmenter} for Transformer-based segmentation, while DEPICT~\cite{wen2024rethinking} and CBT~\cite{wentowards} are two principled improvements upon it.
  }
  % \vspace{-5pt}
  \centering
  \begin{footnotesize}
  \begin{tabular}{lccccc}
    \toprule
    Backbones  
    & Seg. head
    & Resolution
    & FLOPs
    & mIoU (\%)
    \\
    \midrule
    ViT-T & Mask Trans. & 512$^2$ & 13G & 38.1 \\
    ViT-T & DEPICT & 512$^2$ & 13G & 39.3 \\
    ViT-T & CBT & 512$^2$ & \cellcolor{gray!15}12G & \cellcolor{gray!15}39.1 \\
    \rowcolor{cyan!10}
    MiTA-ViT-T$^{\triangledown}$ & CBT & 512$^2$ & 7G{\scriptsize~($\downarrow$42\%)} & 36.5{\scriptsize~(-2.6)} \\
    \midrule
    ViT-S & Mask Trans. & 512$^2$ & 38G & 45.3 \\
    ViT-S & DEPICT & 512$^2$ & 35G & 46.7 \\
    ViT-S & CBT & 512$^2$ & \cellcolor{gray!15}33G & \cellcolor{gray!15}45.8 \\
    \rowcolor{cyan!10}
    MiTA-ViT-S$^{\triangledown}$ & CBT & 512$^2$ & 25G{\scriptsize~($\downarrow$24\%)} & 44.5{\scriptsize~(-1.3)} \\
    \midrule
    ViT-B & Mask Trans. & 512$^2$ & 128G & 48.5 \\
    ViT-B & DEPICT & 512$^2$ & 111G & 49.2 \\
    ViT-B & CBT & 512$^2$ & \cellcolor{gray!15}111G & \cellcolor{gray!15}49.3 \\
    \rowcolor{cyan!10}
    MiTA-ViT-B$^{\triangledown}$ & CBT & 512$^2$ & 95G{\scriptsize~($\downarrow$14\%)} & 48.1{\scriptsize~(-1.2)} \\
    \midrule
    ViT-L & Mask Trans. & 640$^2$ & 667G & 51.8 \\
    ViT-L & DEPICT & 640$^2$ & 628G & 52.9 \\
    ViT-L & CBT & 640$^2$ & \cellcolor{gray!15}622G & \cellcolor{gray!15}53.3 \\
    \rowcolor{cyan!10}
    MiTA-ViT-L$^{\triangledown}$ & CBT & 640$^2$ & 508G{\scriptsize~($\downarrow$18\%)} & 50.5{\scriptsize~(-2.8)} \\    
    \bottomrule
  \end{tabular}
  \label{tab:semantic}
  \end{footnotesize}
  \vspace{-0.5em}
\end{table}
}

\myparagraph{Baselines}
Note that not all methods listed in Tab.~\ref{tab:taxonomy} are included in our experiments for the following reasons:
(i) although representative, many have become outdated; and
(ii) some are designed specifically for language tasks and are therefore not directly comparable in our vision setting. 
Nonetheless, we include comparisons with their vision counterparts that are based on similar ideas.

% \subsection{Classification Evaluated on ImageNet-1K}
\subsection{Image Classification}  %  on ImageNet-1K
Current advances in efficient attention on ImageNet-1K are typically accompanied by extra components such as depth-wise convolutions, 
which obscure the expressiveness gap between the proposed efficient mechanisms and full attention.
To this end, we report a fair comparison in Tab.~\ref{tab:in1k-fair}. 
Specifically, we train all models using the DeiT~\cite{touvron2021training} recipe and vary only the attention mechanism. 
Meanwhile, in Tab.~\ref{tab:in1k-sota}, we compare our best-performing ViT variant based on MiTA
against state-of-the-art and efficient ViTs.

\myparagraph{Implementation details}
% 
% By default, 
We use an image size of $224$ with a patch size of $16$, 
and set $s=1$, $m=k=25$ for MiTA.\footnote{
We set $s=1$ throughout this work 
for a preliminary study.
As analyzed in Sec.~\ref{sec:fit-tax}, this choice does not fully exploit the expressiveness of MiTA.}
Under this setting, each query attends to $m+ks=50$ key-value pairs.
% 
% Moreover, the maximum number of distinct key--value pairs that could be attended (across the $m$ landmark queries, via top-$k$ selection) is upper-bounded by $m\times k = 25^2 = 625$, which is larger than the sequence length $N=197$.
% 

\myparagraph{Results}
As indicated in Tab.~\ref{tab:in1k-fair}, 
without extra components, 
MiTA outperforms other efficient attention by a substantial margin (at least 0.8\% and up to 3.1\%).
Although Agent Attention can narrow the gap by using agent bias~\footnote{
An agent bias $\B \in \R^{N\times m}$ is added to attention scores as $\softmax(\K^\top\tilde{\Q} + \B)$.
}, it still underperforms MiTA on small-sized models by 0.7\%.
Moreover, a severe limitation of this trick is that it is currently not supported by FlashAttention.
As shown in Tab.~\ref{tab:in1k-sota}, when equipped with the architectural modifications explored in ViT-5~\cite{wang2026vit-5}, 
MiTA outperforms a range of strong ViT baselines and approaches state-of-the-art performance within a small margin while using fewer FLOPs.

\subsection{Semantic Segmentation}  %  on ADE20K
ADE20K~\cite{zhou2019semantic} is a scene-centric dataset for semantic segmentation,
which requires higher image resolutions to preserve fine-grained details such as object boundaries.
In Tab.~\ref{tab:semantic}, we apply MiTA to the backbone for efficient visual encoding and compare against methods with ViT backbones and Transformer-based segmentation heads.

\myparagraph{Implementation details}
ViT backbones~\cite{steinertrain} are pretrained on ImageNet-21K with an image size of $384$ and a patch size of $16$.
We use a consistent patch size of $16$ when training on ADE20K, and set $s=1$, $m=k=49$ for MiTA.
Under this setting, 
each query attends to $m+ks=98$ key-value pairs, 
while it is $1{,}024$ (for $512^2$ image resolution) or $1{,}600$ ($640^2$ resolution) in full attention.

\myparagraph{Results}
As depicted in Tab.~\ref{tab:semantic}, MiTA remarkably reduces the FLOPs (by up to 42\%) while achieving comparable segmentation performance. 
Note that MiTA is not fully exploited here because the backbone is not natively pretrained with it.

{

\begin{table}[t]
  \setlength{\tabcolsep}{0.5pt}
  \caption{\small
  Results on LRA benchmark.
  For each task, we report the accuracy and training throughput (steps/s), along with the average accuracy across tasks and the total training wall-clock time (hours).
  $^\ddagger$: a route-only variant (i.e., the compression or local window branch is removed), 
  where the number of entries attended per query kept unchanged by increasing $k$.
  % We report accuracy for each individual task and average accuracy across all tasks.
  % We use $m=16$, $k=16$ for MiTA, %attention, 
  % and $m=32$ for Agent attention.
  % For fair comparison, all methods are implemented \emph{without} DWC. % depth-wise convolutions. 
  % % (DWC).
  }
  % \vspace{-5pt}
  \label{tab:lra}
  \centering
  \begin{footnotesize}
  \begin{tabular}{lllllll}
    \toprule
    Methods & ListOps (2K) & Text (4K) & Retrieval (4K) & Image (1K) & Pathfinder (1K) & Avg. / Tot. \\
    \midrule
    \rowcolor{gray!15}
    Standard Atten & 37.35\,/\,12.7 & 63.63\,/\,3.8 & 79.90\,/\,1.9 & 40.38\,/\,5.7 &  69.68\,/\,5.7 & 58.19\,/\,10.6 \\  % fused, RTX4090
    % Linear Attention & 
    % 18.65\,/\,42.7
    % 65.63\,/\,38.6
    % 80.17\,/\,23.9
    % 38.53\,/\,19.4
    % 72.92\,/\,19.4
    % 55.18\,/\,1.9
    Reformer & 18.55\,/\,34.1 & 65.08\,/\,16.8 & 78.77\,/\,8.5 & 44.04\,/\,8.3 & 69.04\,/\,8.5 & 55.10\,/\,4.5 \\
    Linformer & 38.05\,/\,35.0 & 56.69\,/\,24.6 & 78.41\,/\,13.7 & 39.56\,/\,11.9 & 67.13\,/\,11.9 & 55.97\,/\,3.1 \\
    Performer & 17.84\,/\,28.0 & 65.13\,/\,16.5 & 77.18\,/\,8.9 & 36.64\,/\,8.1 & 70.20\,/\,8.0 & 53.40\,/\,4.7 \\
    Nystr\"omformer & 29.69\,/\,22.0 & 65.84\,/\,16.7 & 79.68\,/\,8.8 & 36.83\,/\,4.5 & 72.50\,/\,4.5 & 56.91\,/\,7.3  \\
    Agent Atten &
    36.59\,/\,40.1&
    63.86\,/\,38.9&
    79.98\,/\,24.3&
    38.92\,/\,19.5&
    71.77\,/\,19.1&
    58.22\,/\,1.9\\
    MoBA$^\ddagger$ & 
    37.10\,/\,30.4&
    62.35\,/\,15.8&
    76.64\,/\,8.2&
    37.92\,/\,8.6&
    69.67\,/\,8.4&
    56.74\,/\,4.6
    \\
    % 256 chunk size, topk 1
    % \\
    % 128 chunk size, top 2
    % 37.40 22.0
    % 62.32 8.8
    % 77.80 4.7
    % 38.78 4.8
    % 70.33 4.8
    % 57.33 8.1
    \rowcolor{cyan!10}
    MiTA$^\ddagger$ & 
    37.20\,/\,25.4 & 
    64.09\,/\,22.2 & 
    79.92\,/\,13.1 & 
    40.17\,/\,5.3 & 
    72.66\,/\,6.3 & 
    58.81\,/\,5.5 \\ % route-only top-256, RTX4090
    \rowcolor{cyan!10}
    MiTA & 
    37.20\,/\,52.5{\scriptsize\,($\times$4.1)} & 
    63.47\,/\,37.7{\scriptsize\,($\times$9.9)} &
    79.88\,/\,19.6{\scriptsize\,($\times$10.3)} &
    40.72\,/\,13.6{\scriptsize\,($\times$2.4)} &
    73.26\,/\,15.3{\scriptsize\,($\times$2.7)} &
    58.91\,/\,2.4{\scriptsize\,($\downarrow$77\%)}
    \\ % 128+128, RTX4090
    \bottomrule
  \end{tabular}
  \end{footnotesize}
  % \vspace{-1.0em}
\end{table}
}

\subsection{Evaluation on Long Sequences}
\begin{wrapfigure}[9]{r}{0.45\textwidth}
\vspace*{-1.2\baselineskip}
  \centering
  % \vspace*{-\baselineskip}
  \includegraphics[width=0.35\columnwidth]{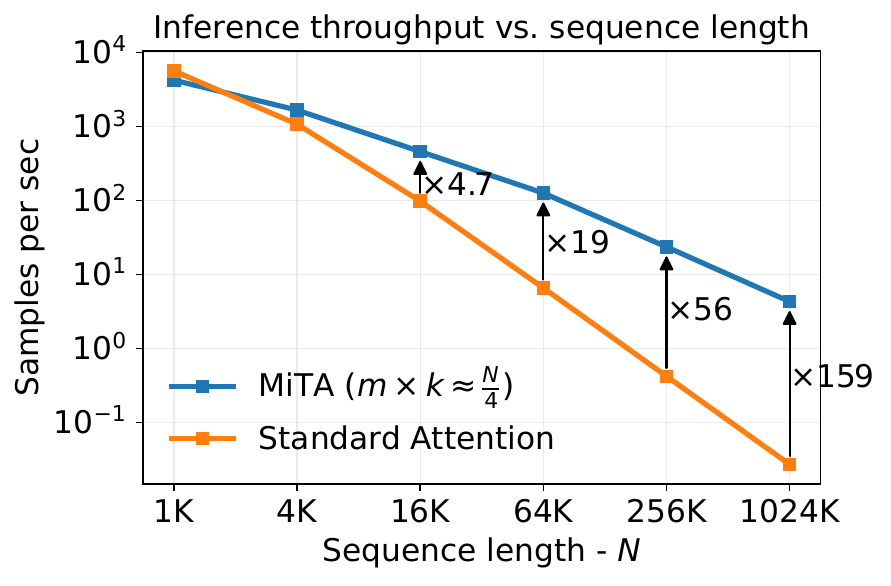}
  \vspace{-1em}
  \vspace{+2pt}
  \caption{\small
    Inference throughput.
    % 
    % The results are measured on a three-layer Transformer with an embedding dimension of 128, 
    % and the batch size is tuned for each run to maximize throughput.    
    % The results are measured on 
    % an NVIDIA RTX 4090 (24GB) with 
    % a three-layer Transformer (model dimension 128) and a synthetic dataset of 10K samples.
    % % 
    % % Best viewed when zoomed in.
  }
  \label{fig:wall-clock}
  % \vspace{-0.5em}
% \end{figure}
\end{wrapfigure}

% Range Arena Benchmark}
As a preliminary validation for long-sequence modeling, we evaluate MiTA on the Long Range Arena (LRA) benchmark~\cite{tay2020long} and compare against several classical efficient attention methods~\cite{kitaev2020reformer, wang2020linformer, choromanski2020rethinking, xiong2021nystromformer}.
We report their accuracy and training throughput (and wall-clock time) on LRA in Tab.~\ref{tab:lra}, 
while the inference throughput improvement of MiTA over standard attention is shown in Fig.~\ref{fig:wall-clock}.

\myparagraph{Implementation details}
The training configuration follows \citet{xiong2021nystromformer} and hyperparameters are kept consistent across all methods for a fair comparison.
Thus, the reported accuracies are not tuned to fully exploit the best performance of each approach.
We set $s=1$ and $m=k=128$ for MiTA.
Throughput and wall-clock time reported in Tab.~\ref{tab:lra} and Fig.~\ref{fig:wall-clock} are measured on a 24GB RTX 4090.
Standard attention is implemented with PyTorch's fused kernel (i.e., FlashAttention). 
More specifically, results in Fig.~\ref{fig:wall-clock} are measured on a three-layer Transformer with an embedding dimension of 128, and the batch size is tuned for each run to maximize throughput.  

% \begin{figure}[h]

\myparagraph{Results}
As shown in Tab.~\ref{tab:lra}, MiTA achieves accuracy comparable to standard attention,
while providing a substantial speedup that reduces total training time by 77\%.
Notably, the route-only variant of MiTA also performs well but is slower, 
since gathering more key-value pairs incurs higher overhead.
This suggest that using a shared, compressed set of key-value pairs 
can effectively represent thousands of routed original pairs.

\section{Conclusion}
We adopted fast-weight scaling as a unifying perspective for efficient attention
% methods 
and introduced a five-dimensional taxonomy to identify the limitations of prior methods. 
%
% With scaling strategy as the most central axis of this taxonomy, 
Moreover, we proposed a novel efficient attention method, 
% which is termed as
% the mixture of top-$k$ activations (MiTA),
termed Mixture-of-Top-$k$ Attention (MiTA),
%{\mcr
which constructs a tunable number of query-aware and deformable fast-weight experts and 
bridges the two 
% previously isolated approaches---scaling by routing and scaling by compression 
different scaling strategies together, thereby addressing their inherent limitations.
% ---scaling by routing and scaling by compression---and 
%}
%
% Beyond this, MiTA attention also instantiates and enriches the other four dimensions of the taxonomy.
%
% While our empirical validation is preliminary, 
% 
We provided a detailed analysis of both the taxonomy and our MiTA, suggesting a principled and promising path for developing efficient attention from the fast-weight scaling perspective.

% \section*{Limitations} \label{sec:limit}
% % 
% This work primarily focuses on the vision domain, and we do not evaluate the proposed method in the decoding phase of LLMs. This choice is mainly due to our limited computational resources. 
% % 
% % That said,
% Nonetheless,
% the method and related work are presented in a modality-agnostic manner, and we discuss their applicability beyond vision. We consider the application to LLMs as an important direction for future work. The main contribution of this paper is conceptual, namely a unified perspective that facilitates principled improvements, rather than a domain-specific instantiation.

% \begin{ack}
% Use unnumbered first level headings for the acknowledgments. All acknowledgments
% go at the end of the paper before the list of references. Moreover, you are required to declare
% funding (financial activities supporting the submitted work) and competing interests (related financial activities outside the submitted work).
% More information about this disclosure can be found at: \url{https://neurips.cc/Conferences/2026/PaperInformation/FundingDisclosure}.

% Do {\bf not} include this section in the anonymized submission, only in the final paper. You can use the \texttt{ack} environment provided in the style file to automatically hide this section in the anonymized submission.
% \end{ack}

% \section*{References}

{
\small
% \bibliographystyle{plainnat}
% \bibliographystyle{unsrtnat}
% \bibliography{nips26}

}

%%%%%%%%%%%%%%%%%%%%%%%%%%%%%%%%%%%%%%%%%%%%%%%%%%%%%%%%%%%%

\clearpage
\appendix
\begin{center}
    \LARGE \textbf{Appendix}
\end{center}

\section{Ablation Study}\label{sec:ablation}

In Tab.~\ref{tab:unified_ablation}, we ablate three key design choices in MiTA:
1) the landmark extraction strategy;
2) the number of experts ($m$) and the number of key-value pairs per expert ($k$); and
3) the standalone effects of routing and compression.
In the following paragraphs, we analyze these results and their implications.

\begin{table}[htbp]
  \vspace{-10pt}
  \caption{Ablation study under the same setting as Tab.~\ref{tab:in1k-fair}.}
  \label{tab:unified_ablation}
  \centering
  \footnotesize
  \begin{tabular}{lcc}
    \toprule
    Setting & Acc. & $\Delta$ \\
    \midrule
    \multicolumn{3}{c}{Landmark Extraction} \\
    \midrule
    Random Selection & 70.6 & -0.5 \\
    Learnable Parameters & 66.3 & -4.8 \\
    1D Average Pooling & 70.4 & -0.7\\
    % \rowcolor{gray!15}
    \rowcolor{cyan!10}
    2D Average Pooling & 71.1 & Default\\
    Convolution & 67.0 & -4.1 \\
    Depth-Wise Convolution & 68.8 & -2.3 \\
    Token Merging~\cite{boly:tome} & \textbf{71.3} & \textbf{+0.2} \\
    % \addlinespace[2pt]
    \midrule
    \multicolumn{3}{c}{$m \times k$} \\
    \midrule
    16 $\times$ 16 & 70.0 & -1.1 \\
    16 $\times$ 25 & 71.0 & -0.1 \\
    25 $\times$ 16 & 70.5 & -0.6 \\
    \rowcolor{cyan!10} 
    25 $\times$ 25 & 71.1 & Default \\
    25 $\times$ 36 & 71.7 & +0.6 \\
    36 $\times$ 25 & 71·2 & +0.1 \\
    36 $\times$ 36 & 71.7 & +0.6 \\
    % \addlinespace[2pt]
    \midrule
    \multicolumn{3}{c}{Compression \& Routing} \\
    \midrule
    \rowcolor{cyan!10}
    Compress-and-route & 71.1 & Default \\
    Compress-only & 70.3 & -0.8 \\
    Route-only & 70.8 & -0.3 \\
    \bottomrule
  \end{tabular}
\end{table}

\begin{figure}[htbp]
  \centering
  \includegraphics[width=0.45\columnwidth]{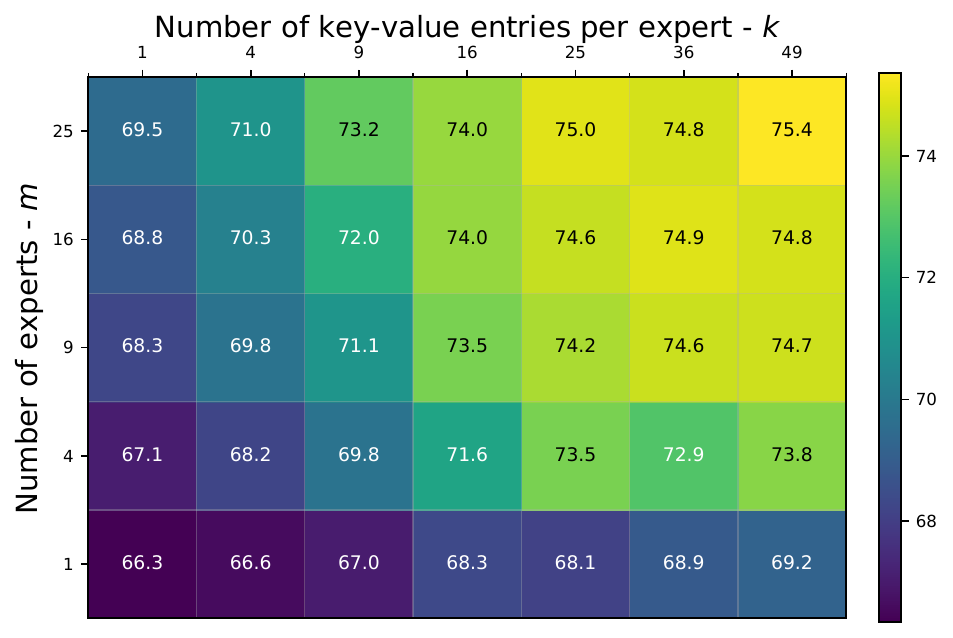}
  \vspace{-6pt}
  \caption{Ablation study of $m$ and $k$ on CIFAR-100.
  Experiments with other $(m, k)$ were not conducted due to out-of-memory (OOM).
  }
  \label{fig:ablation-cifar}
\end{figure}

\myparagraph{Landmark extraction} 
Our simple default choice --- average pooling over evenly split, non-overlapping rectangular regions --- even outperforms parameterized alternatives.
This may stem from the fact that landmark queries serve the dual roles of compressed keys and routers, potentially leading to gradient conflicts and suggesting a direction for further improving MiTA.
We also observe that more advanced methods, 
such as ToMe~\cite{boly:tome} can improve MiTA. 
However, ToMe must be applied repeatedly to achieve aggressive compression,
making it a less lightweight and less elegant solution. 
% for attaining arbitrary compression ratios.

\myparagraph{Hyperparameters $m$ and $k$}
Performance improves as either $m$ or $k$ increases, which is expected since MiTA more closely approximates full attention with larger $m$ or $k$, and recovers full attention when $m=k=N$.
More importantly, we observe that performance is more sensitive to $k$ than to $m$; that is, increasing $k$ is typically more beneficial than increasing $m$. 
This trend is further validated in Fig.~\ref{fig:ablation-cifar}.
This finding supports our motivation that gathering top-$k$ activations independently for each query (e.g., as in Spark Attention~\cite{you2025spark} and DeepSeek Sparse Attention~\cite{liu2025deepseek}) is redundant, and that many such activations can instead be shared through routing.
For practical use, we provide a simple rule of thumb: first choose a fixed ratio, $\frac{m \times k}{N}$; then start from $m=k$ and explore $k > m$ during subsequent tuning.

\myparagraph{Compression and routing}
Routing is more critical than compression. Nevertheless, introducing compression further improves performance while also providing notable acceleration, as shown in Tab.~\ref{tab:lra}.

\begin{figure}[htbp]
  \centering
  \includegraphics[width=\columnwidth]{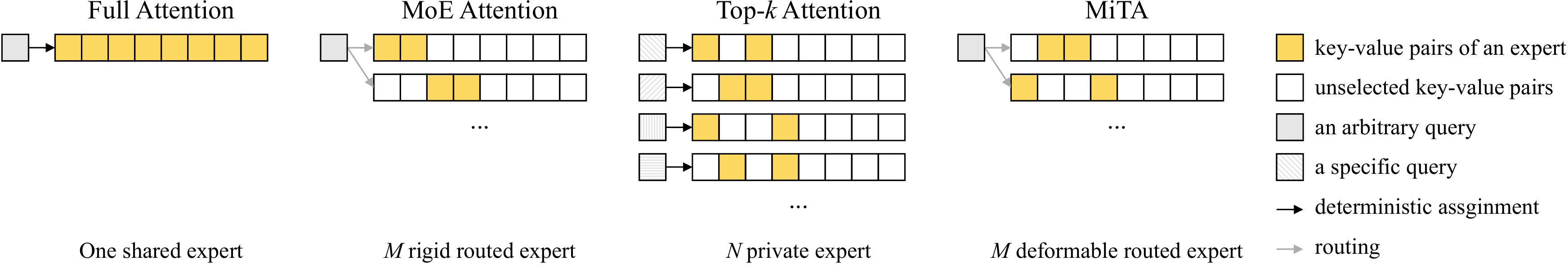}
  \caption{
    Visual comparison of related attention methods.
  }
  \label{fig:attention-compare}
\end{figure}

\section{Fit MiTA into the Fast-Weight Scaling Taxonomy} \label{sec:fit-tax}
% 
% While MiTA is introduced from the angle to combine the two complementary scaling strategies in Sec.~\ref{sec:mita},
While MiTA is introduced in Sec.~\ref{sec:mita} from the perspective of combining the two complementary scaling strategies,
% scaling by routing and scaling by compressing.
% 
% Moreover, 
we find that it also 
enriches
% refreshes 
% the approaches in
the remaining
% other four 
dimensions of the
fast-weight scaling taxonomy proposed in Sec.~\ref{sec:fast-weight-scaling}. 
% e.g., by constructing deformable fast-weight experts.
% 
% In this subsection, we briefly discuss each of them.
This subsection will briefly discuss each of them.

% \subsubsection{Expert Construction}
\myparagraph{Expert construction}
% 
% How 
% fast-weight 
% experts 
% are constructed
% naturally
% Expert construction T
This core dimension
determines 
the expert type and count, and hence the routing topology.
% 
% We argue that
% 
An important consideration 
% in it
% expert construction
is what the construction conditions on:
the query only (e.g., DAT~\cite{xia2022vision}), 
the key-value pairs only (e.g., MHLA~\cite{zhang2026mhla}), or both.
When 
% conditioning on 
using
key-value pairs,
the construction may be content-dependent and thus deformable (e.g., DSA~\cite{liu2025deepseek}) or 
merely position-driven (e.g., MoBA~\cite{Lu:2025-MoBA}).
While most efficient attention methods condition on one of the above aspects,
our MiTA
% like full attention, 
conditions on all of them 
by extracting landmark queries from queries and 
then probe the full key–value pool.
We illustrate a resulting benefit of this design in Fig.~\ref{fig:token-pruning}.

% Although landmark queries can be made query-agnostic
% % (i.e., slow-weight),
% % , e.g., 
% by setting them as learnable parameters (slow weights), 
% this is not recommended.
% %
% \citet[Table~17]{han2024agent} ablate different landmark query extraction methods 
% and find that average pooling performs best.
% %
% % Moreover, 
% Furthermore, \citet[Appendix~A]{Wen:NIPS25-CBSA} find that slow-weight landmark queries render
% $\softmax(\K^\top \tilde{\Q})$ low-rank, thereby reducing the effective number of landmark queries
% and leading to inferior performance.

% \subsubsection{Expert Type and Count}
\myparagraph{Expert type and count}
Expert types can be categorized in three classes:
a) 
% restricted to 
linear layers (e.g., linear attention),
b)
% restricted to 
MLPs (e.g., sparse attention and PVT~\cite{Wang2021pvt}),
c) 
% applicable to 
arbitrary modules (e.g., test-time training).
The scaling-by-routing branch of MiTA is restricted to MLP experts, as in sparse attention.
However, the scaling-by-compression branch
can be implemented through test-time training, thereby generalizing 
% the shared expert $\bar{\mathcal{E}}$ in Eq.~\eqref{eq:shared-expert}
to other modules.
As for the expert count,
MiTA controls it via the
% tunable 
hyperparameter $m$. 
One benefit of this design has been discussed in 
% the 
complexity analysis. 
Another benefit is that it enables expert composition through the routing mechanism.
% routing to play a non-trivial role.

% can draw inspiration from test-time training
% and optimization unrolling~\cite{monga2021algorithm}.
%  gregor2010learning, 

\begin{figure}[h]
  \centering
  \hspace{-10pt}
  \includegraphics[width=0.48\columnwidth]{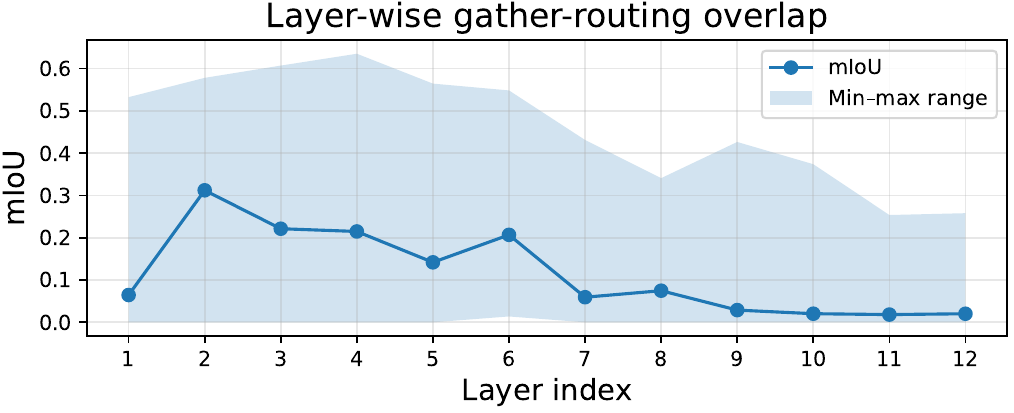}
  \vspace{-6pt}
  \caption{
  % We quantify 
  The layer-wise positional overlap 
  (between the key–value pairs gathered by an expert and the queries routed to it)
  is quantified by mIoU, 
  % using mean intersection-over-union (mIoU), 
  averaged over experts and attention heads.
  }
  \label{fig:overlap}
\end{figure}

\myparagraph{Routing topology}
In MiTA, there is a fixed number 
% ($m$) 
of base sparse patterns (i.e., fast-weight experts),
which is a substantial improvement over efficient attention with only a single expert,
yet is still far fewer than prior top-$k$ attention~\cite{you2025spark, liu2025deepseek}, which has $N$ experts.
Crucially, to compensate for this limitation,
MiTA 
% relies on 
can resort to the routing mechanism to compose these $m$ base experts:
when routing each query to $s$ experts, the number of effective sparse patterns (i.e., the %combination 
combinatorial number of experts)
is $\binom{m}{s}$.
% = \nicefrac{m!}{s!(m-s)!}$.
% 
% As a preliminary study on a classical vision task (e.g., image classification),
% we implement MiTA with $s=1$ throughout.
% 
% In particular, 
In particular, 
when $s=1$, 
routing may degenerate into clustering, 
restricting information flow within each cluster.
% 
% To assess this risk, 
We thus quantify the positional overlap between an expert’s gathered key–value pairs 
and the queries routed to it. 
As shown in Fig.~\ref{fig:overlap}, 
the overlap remains consistently modest across layers, 
suggesting that MiTA with $s=1$ performs routing rather than hard clustering.

\vspace{-6pt}

{
\setlength{\textfloatsep}{8pt plus 2pt minus 2pt}
\begin{algorithm}[t!]
% \small
  \caption{Mixture-of-Top-$k$ Attention (MiTA)}
  \label{alg:mita}
  {\setlength{\baselineskip}{1.12\baselineskip}
  \begin{algorithmic}[1]
    \REQUIRE
    Query, key, value matrices $\Q, \K, \V \in \R^{d\times N}$;
    the number of landmark queries $m$;
    the width of each fast-weight expert $k$;
    each query is routed to the shared expert and to $s=1$ additional expert.
    \STATE \texttt{// Obtain landmark queries}
    \STATE $\tilde{\Q} = \operatorname{AdaptiveAvgPool}(\Q, \textrm{output\_size=}m)$ 
    \hfill \texttt{// [d, m]}
    \STATE \texttt{// Lookup key-value pairs via landmark queries}
    \STATE $\S^\mathrm{kv} = \K^\top\tilde{\Q} / \sqrt{d}$
    \hfill \texttt{// [N, m]}
    \STATE \texttt{// Gather the top-$k$ activated key-value pairs}
    \STATE $\mathcal{I}^\mathrm{kv} = \operatorname{Flatten}(\operatorname{TopK}(\smash{\S^\mathrm{kv}}^\top, k, \textrm{dim=}1))$
    \hfill \texttt{// [m*k]}
    \STATE $\K^\mathrm{expt},\V^\mathrm{expt} = \K[:, \mathcal{I}^\mathrm{kv}], \V[:, \mathcal{I}^\mathrm{kv}]$
    \hfill \texttt{// [d, m*k]}
    \STATE \texttt{// Obtain landmark values (construct the shared expert)}
    \STATE
    $\tilde{\V} = \V \softmax(\mathbf{S}^{\mathrm{kv}})$
    \hfill \texttt{// [d, m]}
    \STATE \texttt{// Always route queries to the shared expert}
    \STATE $\O^\mathrm{share} = \operatorname{FlashAttention}(\Q, \tilde{\Q}, \tilde{\V})$
    \STATE \texttt{// Sparsely route queries to other experts}
    \STATE $\mathcal{I}^\mathrm{expt} = \operatorname{ArgSort}(\operatorname{ArgMax}(\smash{\tilde{\Q}}^\top \Q, \textrm{dim=}0))$
    \hfill \texttt{// [N]}
    \STATE
    $\O^\mathrm{expt} = \operatorname{FlashAttention}(\Q[:, \mathcal{I}^\mathrm{expt}], \K^\mathrm{expt}, \V^\mathrm{expt},$ \\
    $\textrm{cu\_seqlens\_q=}\operatorname{CumSum}(\operatorname{BinCount}(\mathcal{I}^\mathrm{expt})),$
    $\textrm{cu\_seqlens\_k=}[0, k, 2k, \ldots, mk])$
    \STATE \texttt{// Combine results via online softmax}
    \STATE
    $\O = \operatorname{Combine}(\O^\mathrm{share}, \O^\mathrm{expt})$
    \hfill \texttt{// [d, N]}
    \STATE \textbf{return} $\O$
  \end{algorithmic}
  }
\end{algorithm}
}

\begin{figure}[h]
  \centering
  % \hfill
  \hspace{-2pt}
  % 
  %   \begin{minipage}[b]{0.24\columnwidth}
  %   \centering
  %   \includegraphics[width=\linewidth]{hyperparameter_generalization_finetune_tiny_512.pdf}
  %   \caption*{\scriptsize (a) DeiT-T models}
  % \end{minipage}
  % % 
  %   \begin{minipage}[b]{0.24\columnwidth}
  %   \centering
  %   \includegraphics[width=\linewidth]{hyperparameter_generalization_finetune_large.pdf}
  %   \caption*{\scriptsize (a) DeiT-T models}
  % \end{minipage}
  % 
  \begin{minipage}[b]{0.3\columnwidth}
    \centering
    \includegraphics[width=\linewidth]{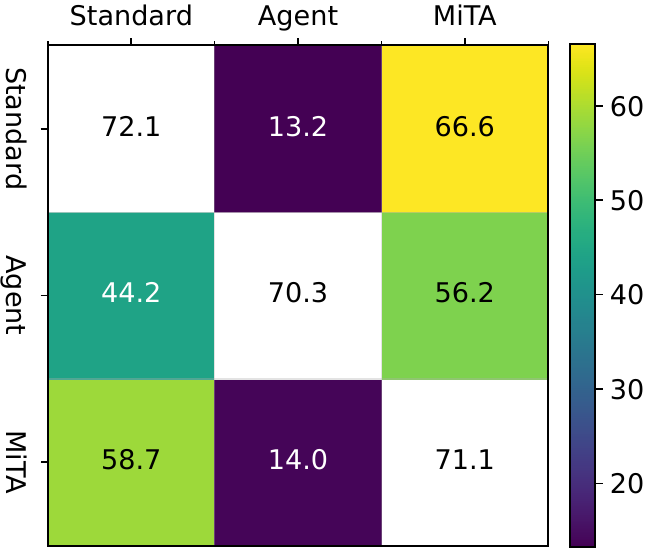}
    \caption*{\scriptsize (a) DeiT-T models}
  \end{minipage}
  % \hfill
  \hspace{8pt}
  \begin{minipage}[b]{0.3\columnwidth}
    \centering
    \includegraphics[width=\linewidth]{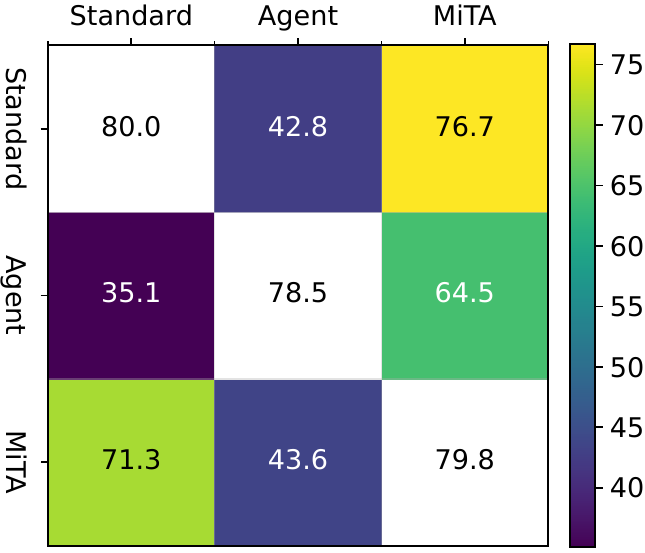}
    \caption*{\scriptsize (b) DeiT-S models}
  \end{minipage}
  % \hfill
  \caption{
    % Checkpoint swapping.
    Evaluation with a different inference attention. 
    The x-axis indexes the training attention,
    while the y-axis indexes the inference attention.
    % 
    % The y-axis indexes the target model 
    % (i.e., the attention mechanism used at inference time), 
    % while the x-axis indexes the source weights 
    % (i.e., the attention mechanism under which the weights were trained).
    % 
    We omit the diagonal entries from the heatmap since they are not of interest.
    % 
    % All models are trained from scratch on ImageNet-1K.
    Currently, only standard attention, Agent Attention, and MiTA are included.
    % Best viewed when zoomed in.
  }
  % \vspace{-2em}
  \label{fig:attn-gen}
\end{figure}

\begin{figure}[h]
  \centering
  % \hfill
  % \hspace{-20pt}
  \begin{minipage}[b]{0.36\columnwidth}
    \centering
    \includegraphics[width=\linewidth]{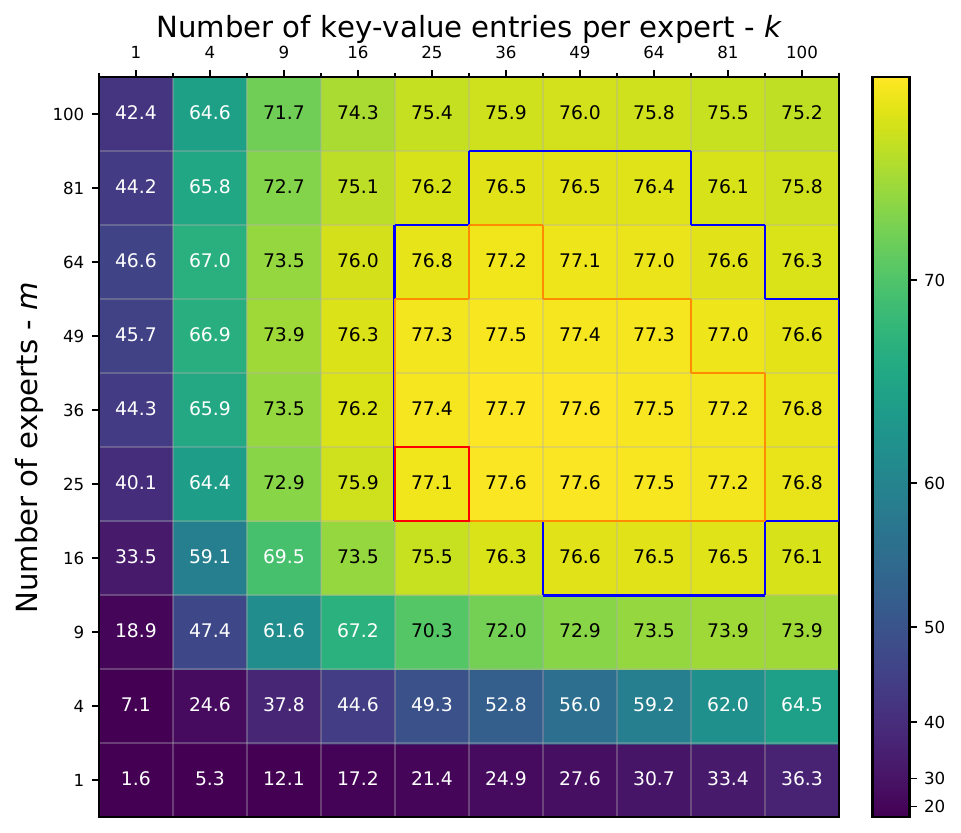}
    \caption*{\scriptsize (a) MiTA-ViT-T @$512$}
  \end{minipage}
  % \hfill
  \hspace{8pt}
  \begin{minipage}[b]{0.36\columnwidth}
    \centering
    \includegraphics[width=\linewidth]{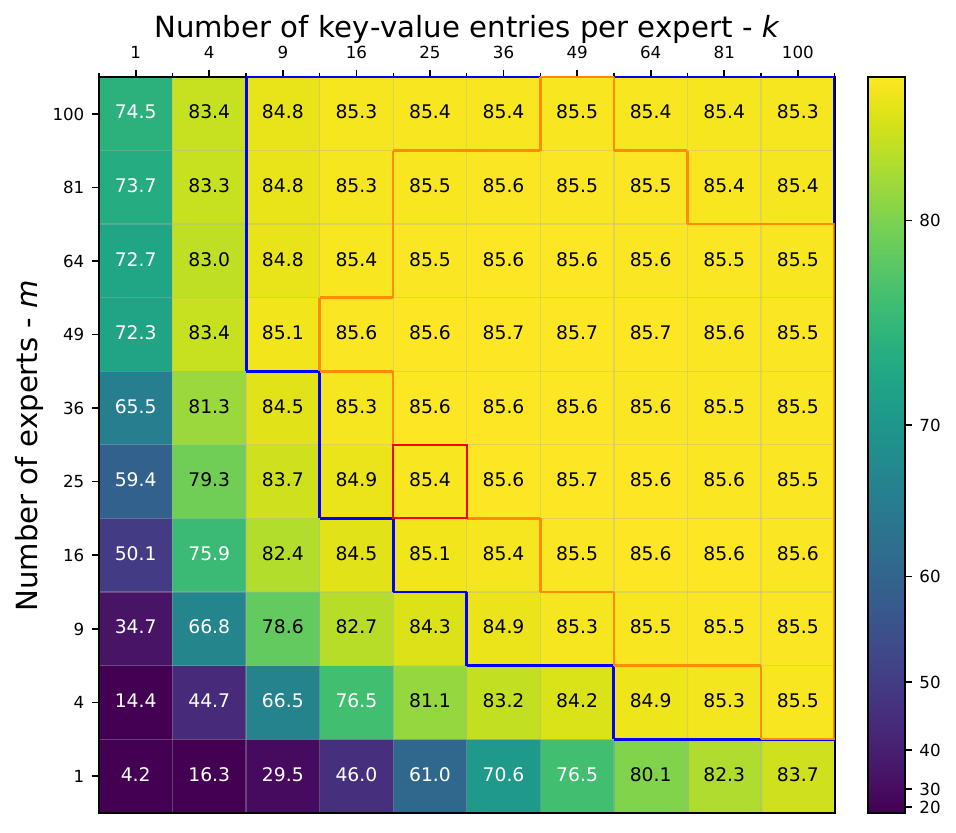}
    \caption*{\scriptsize (b) MiTA-ViT-L @$224$}
  \end{minipage}
  % \hfill
  \caption{
    MiTA's generalization across $m$ and $k$.
    The red box marks the training (or finetuning) $(m, k)$ and the corresponding baseline accuracy on ImageNet-1K.
    The orange boundary marks the inference $(m, k)$ configurations that exceed the baseline accuracy, and
    the blue boundary marks those achieve 99\% of the baseline accuracy.
    Best viewed when zoomed in.
  }
  % \vspace{-3pt}
  \label{fig:m-k-gen}
\end{figure}

\clearpage

\section{Algorithmic Generalization} \label{sec:generalization}
Broadly, generalization typically refers to performance beyond the training data, 
e.g., 
from the training set to the validation set, 
length extrapolation~\cite{zhao2024length} (from the training resolution to a new one). 
% or transfer learning (from one dataset to another).
% 
Motivated by the unifying perspective for efficient attention, 
this work investigate a new dimension of generalization: \emph{algorithmic generalization}, i.e., 
performance beyond the training algorithm (e.g., attention mechanism).
% \footnote{
% As in typical generalization, the model parameters are fixed; only the algorithm changes.
% }

In Fig.~\ref{fig:m-k-gen}, we study MiTA's generalization across hyperparameters:
the expert count \(m\) and expert width \(k\).
Note that changing an algorithm's hyperparameters is analogous to changing the input resolution.
Moreover, in Fig.~\ref{fig:attn-gen}, we study how an attention mechanism generalizes to others.
Specifically, given a model trained with one attention mechanism (\emph{training attention}), we replace it with a different mechanism (\emph{inference attention}) at test time.
Beyond the generalization setting, where model parameters are fixed,
we also finetune models pretrained on a large-scale dataset with a different \emph{finetuning attention} in Tab.~\ref{tab:in21k-ft}.

\myparagraph{Implementation details}
ViTs~\cite{steinertrain} in Tab.~\ref{tab:semantic} are pretrained on ImageNet-21K with an image size of $224$ and a patch size of $16$.
We finetune them for 50 epochs with the AdamW optimizer, using the listed attention mechanisms (including standard attention).
% Models in Tab.~\ref{tab:semantic} and Fig.~\ref{fig:m-k-gen} are fine-tuned at an image size of $224$, except for Fig.~\ref{fig:m-k-gen}(a).
The model in Fig.~\ref{fig:m-k-gen}(a) is similarly finetuned, but starts from a ViT pretrained with an image size of $384$.
Models in Fig.~\ref{fig:attn-gen} are the same as those in Tab.~\ref{tab:in1k-fair}.

\myparagraph{Results}
Fig.~\ref{fig:m-k-gen} suggests that MiTA generalizes well when scaling up the expert count and width: many configurations exceeds the original result.
This indicates a promising strategy: train MiTA with smaller \(m\) and \(k\) for efficiency, then increase \(m\) or \(k\) at inference to obtain gains.
In Fig.~\ref{fig:attn-gen}, we observe that standard attention and MiTA generalize to each other better than other pairs, 
especially when training with standard attention and testing with MiTA, 
which retains over 95\% of the original performance with linear complexity.
Regarding finetuning, Tab.~\ref{tab:in21k-ft} shows that 
parameters pretrained with standard attention 
transfer more easily to MiTA,
while Agent Attention struggles to narrow this gap 
even when increasing the number of agent tokens. 

\begin{table}[h]
\setlength{\tabcolsep}{4pt}
  \caption{
    Finetuning results.
    We finetune ImageNet-21K pretrained ViTs~\cite{steinertrain} on ImageNet-1K, replacing the standard attention with Agent Attention or MiTA Attention.
    $^\dagger$: the number of agent tokens (i.e., landmark queries) is increased from 49 (default) to 64.
    }
%   ImageNet-1K finetune results.
%   ViTs are pretrained on ImageNet-21K, and then finetuned with the attention mechanism replaced.
%   We finetune all models for 50 epochs. % to ensure sufficient convergence.
% % All models are trained and evaluated at an image resolution of $224\times224$ with a patch size of $16\times16$.
  \label{tab:in21k-ft}
  \centering
  \begin{footnotesize}
  \begin{tabular}{lcccc}
    \toprule
    Methods & Tiny & Small & Base & Large \\
    \midrule
    \rowcolor{gray!15}
    Standard Attention & 76.9 & 81.2 & 84.4 & 85.9 \\
    Linear Attention & 73.2 & 78.6 & 79.8 & 80.7 \\
    Agent Attention & 74.6 & 79.7 & 81.4 & 83.5
    \\
    Agent Attention$^\dagger$ & 74.7 & 79.9 & 81.5 & 83.8
    \\
    \rowcolor{cyan!10}
    MiTA Attention & 
    75.6{\scriptsize\,(-1.3)} & 
    80.9{\scriptsize\,(-0.3)} & 
    82.8{\scriptsize\,(-1.6)} & 
    85.5{\scriptsize\,(-0.4)} \\
    \bottomrule
  \end{tabular}
  \end{footnotesize}
\end{table}

\section{Limitations} \label{sec:limit}
This work primarily focuses on the vision domain, and we do not evaluate the proposed method in the decoding phase of LLMs. This choice is mainly due to our limited computational resources. 
That said, the method and related work are presented in a modality-agnostic manner, and we discuss their applicability beyond vision. We consider the application to LLMs as an important direction for future work. The main contribution of this paper is conceptual, namely a unified perspective that facilitates principled improvements, rather than a domain-specific instantiation.

%%%%%%%%%%%%%%%%%%%%%%%%%%%%%%%%%%%%%%%%%%%%%%%%%%%%%%%%%%%%
% \clearpage
% \newpage
% \input{checklist.tex}

\end{document}